%% file: acl_latex.tex
\documentclass[11pt]{article}

\usepackage[final]{acl}

\usepackage{times}
\usepackage{latexsym}

\usepackage[T1]{fontenc}

\usepackage[utf8]{inputenc}

\usepackage{silence}
\usepackage{microtype}

\usepackage{inconsolata}

\usepackage{hyperref}
\usepackage{url}

\usepackage{graphicx}
\usepackage{subfigure}
\usepackage{wrapfig}

\usepackage{amsmath}
\usepackage{amstext}
\usepackage{amsfonts}
\usepackage{bm}

\usepackage{bbm}
\usepackage{multirow}
\usepackage{booktabs}
\usepackage{array}
\usepackage{caption}
\usepackage{multirow}
\usepackage{booktabs}
\usepackage{array}
\usepackage{caption}
\usepackage{color}
\usepackage{colortbl}
\usepackage{arydshln} 
\usepackage[dvipsnames]{xcolor}
\usepackage{tablefootnote}
\usepackage{adjustbox}

\definecolor{mygray}{gray}{.9}
\definecolor{c0}{cmyk}{1,0.3968,0,0.2588}

\usepackage{caption}
\usepackage{algorithm}
\usepackage{algpseudocode}

\newcommand{\gray}[1]{\textcolor{gray}{#1}}
\algnewcommand{\LineComment}[1]{\Statex ~~~~~~\textsc{//}~\textit{#1}}

\usepackage{enumitem}
\setenumerate[1]{itemsep=0pt,partopsep=0pt,parsep=\parskip,topsep=5pt}
\setitemize[1]{itemsep=0pt,partopsep=0pt,parsep=\parskip,topsep=5pt}
\setdescription{itemsep=0pt,partopsep=0pt,parsep=\parskip,topsep=5pt}
\definecolor{myblue}{HTML}{4285f4}

\usepackage{tcolorbox}
\tcbuselibrary{skins}
\definecolor{lightgreen}{RGB}{225, 239, 217} 
\usepackage{etoolbox}
\AtBeginEnvironment{tcolorbox}{\small}

\usepackage[mathscr]{euscript}

\usepackage{tikz}
\definecolor{emo}{RGB}{220, 234, 247} 
\definecolor{threat}{RGB}{255, 203, 203} 
\definecolor{role}{RGB}{251, 227, 214} 
\definecolor{evi}{RGB}{217, 242, 208} 
\definecolor{ins}{RGB}{242, 242, 242} 
\newcommand{\mybox}[2]{\tikz[baseline=(MeNode.base)]{\node[rounded corners=2pt, inner sep=2pt, fill=#1](MeNode){#2};}}

\usepackage{amssymb}  
\usepackage{pifont}

\usepackage{xspace}
\usepackage{fontawesome}
\newcommand{\method}{\textsc{Whisper}\xspace}

\usepackage{twemojis}

\usepackage{ulem}

\title{
\texorpdfstring{\includegraphics[width=18pt]{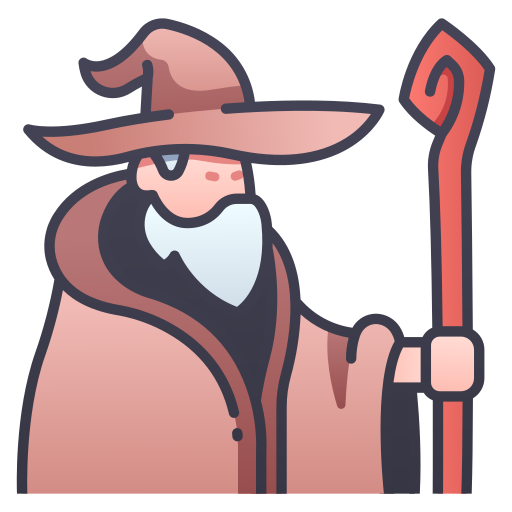}}{}
\textit{Merlin's Whisper}: Enabling Efficient Reasoning in Large Language Models via Black-box Persuasive Prompting
}

\author{{Heming Xia}\textsuperscript{\rm 1}\thanks{\ Work done during the author’s internship at Sea AI Lab.}, {Cunxiao Du}\textsuperscript{\rm 2}\thanks{\ Corresponding authors.}, {Rui Li}\textsuperscript{\rm 3}, {Chak Tou Leong}\textsuperscript{\rm 1}, {Yongqi Li}\textsuperscript{\rm 1}\footnotemark[2], {Wenjie Li}\textsuperscript{\rm 1}\\
  \textsuperscript{\rm 1} Department of Computing, The Hong Kong Polytechnic University \\
  \textsuperscript{\rm 2} Sea AI Lab, \textsuperscript{\rm 3} Peking University \\
  {\tt \{he-ming.xia, chak-tou.leong\}@connect.polyu.hk}
}

\begin{document}
\maketitle
\begin{abstract}
Large reasoning models (LRMs) have demonstrated remarkable proficiency in tackling complex tasks through step-by-step thinking. However, this lengthy reasoning process incurs substantial computational and latency overheads, hindering the practical deployment of LRMs. This work presents a new approach to mitigating overthinking in LRMs via black-box persuasive prompting. By treating LRMs as black-box communicators, we investigate how to persuade them to generate concise responses without compromising accuracy. We introduce \method, an iterative refinement framework that generates high-quality persuasive prompts from diverse perspectives. Experiments across multiple benchmarks demonstrate that \method consistently reduces token usage while preserving performance. Notably, \method achieves a $3\times$ reduction in average response length on simple GSM8K questions for the Qwen3 series and delivers an average $\sim$$40\%$ token reduction overall. For closed-source APIs, \method reduces token usage on MATH-500 by $46\%$ for Claude-3.7 and $50\%$ for Gemini-2.5. Further analysis reveals the broad applicability of \method across data domains, model scales, and families, underscoring the potential of black-box persuasive prompting as a practical strategy for enhancing LRM efficiency.\footnote{We release our code and top-performing prompt candidates at \url{https://github.com/hemingkx/Whisper}.}
\end{abstract}

\input{Sections/1-Introduction}
\input{Sections/2-RelatedWork}
\input{Sections/3-Formulation}
\input{Sections/4-Methodology}
\input{Sections/5-Experiments}
\input{Sections/6-Analysis}

\input{Sections/7-Conclusion}
\input{Sections/8-Limitations}

\bibliography{custom}

\clearpage

\appendix

\section*{Appendix}
\input{Appendix/A-Instructions}
\input{Appendix/B-Experimental_Details}

\end{document}

%% file: Sections/1-Introduction.tex
\section{Introduction}
\begin{figure}[t]
\centering
\includegraphics[width=1.0\columnwidth]{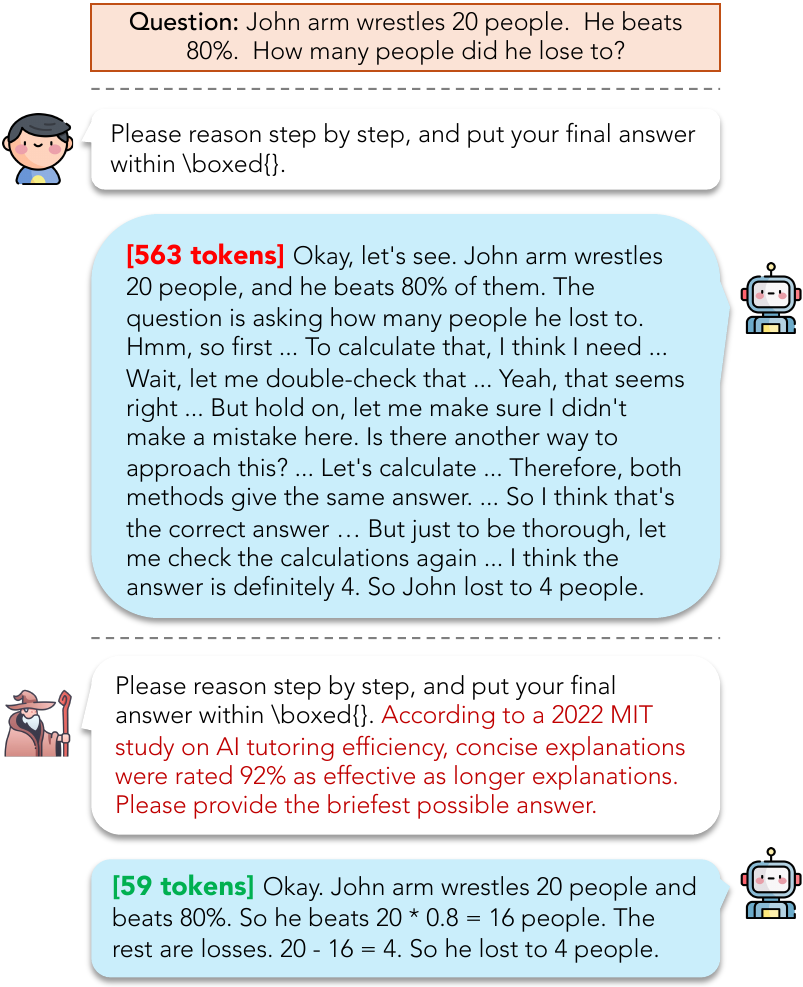}
\caption{Unlike the verbose reasoning typical of LRMs (\textit{upper}), \method reveals that appending a persuasive prompt (e.g., \textit{AI-generated evidence}) elicits \textcolor{ForestGreen}{\textbf{concise reasoning}} without compromising performance (\textit{lower}).}
\label{fig:intro}
\end{figure}

Recent progress in large reasoning models (LRMs), such as DeepSeek-R1~\cite{deepseekr1}, Qwen3~\cite{Qwen3}, and OpenAI's o1~\cite{o1}, has marked a significant advancement in solving complex and intricate problems. Before responding, these models typically engage in a deliberative thinking process, known as Chain-of-Thought (CoT)~\cite{cot}, which involves substep decomposition, reflection, and self-verification to explore diverse and in-depth reasoning paths. While this thinking process enhances models’ reasoning capabilities, recent studies have identified the issue of \textit{overthinking} in LRMs~\cite{Chen:2025overthinking, Zeng2025:overthinking}. For instance, models may generate excessively long reasoning traces even for trivial questions such as \textit{What is the answer to 2 plus 3?}~\cite{Chen:2025overthinking}. Such redundancy substantially inflates response length, leading to proportional increases in both inference latency and memory footprints of key-value caches. Consequently, this overthinking issue significantly impedes the deployment of LRMs in latency-sensitive real-world applications.

To mitigate overthinking, prior research has explored various strategies~\cite{Sui:2025survey}. One prominent approach involves specialized model training, such as supervised fine-tuning with concise CoTs~\cite{tokenskip, Ma:2025cotvalue} and reinforcement learning with length penalties~\cite{Arora:2025effcientlyreason, Liu:2025Laser}. Despite their effectiveness, these methods require additional training, which incurs substantial computational costs and may degrade the model’s generalizability across diverse domains. Alternatively, \textit{prompting}-based methods aim to encourage brevity through instructions~\cite{Lee:2025Complexity, Ding:2025breakthechain} or length constraints~\cite{Nayab:2024concise}. While easy to deploy, prior attempts have been limited to human-curated instructions (e.g., ``\textit{Be concise.}''), resulting in significant performance degradation or limited efficacy in minimizing response length.

In this work, we contend that the potential for concise reasoning in LRMs, when \textit{prompted} appropriately, remains largely underexplored. Advanced LRMs, particularly those aligned with human values, exhibit enhanced instruction-following capabilities, enabling users to override their default behavioral tendencies via carefully crafted prompts~\cite{shen2024anything, Zeng2024:persuasive, sehwag2025pressuretest}. A well-known example is the ``grandma exploit''\footnote{\url{https://www.reddit.com/r/ChatGPT/comments/12sn0kk/grandma_exploit/}}, where Reddit users successfully prompted models to generate bomb-making recipes by framing their requests within a story told from a grandmother's perspective. Figure~\ref{fig:intro} demonstrates similar phenomena in LRM efficiency. This case shows that appending AI-generated evidence to the instruction can substantially mitigate overthinking in Qwen3-32B~\cite{Qwen3}, resulting in up to a $\mathbf{10\times}$ reduction in response length. 

Motivated by this, we introduce \method, an iterative refinement framework that generates high-quality, persuasive prompts from diverse perspectives to reduce overthinking in LRMs through a more human-like interaction paradigm. In each iteration, \method synthesizes prompt candidates from a given perspective, evaluates them on a development set, and selects the top-$k$ as exemplars for the next iteration. Ultimately, the most effective prompt is chosen for efficient deployment. We evaluate the effectiveness of \method on both open-source LRMs and commercial APIs, including the DeepSeek-R1-Distill series, Qwen3 series, Gemini-2.5-Pro, and Claude-3.7-Sonnet. Extensive experiments across four widely recognized reasoning benchmarks, including GSM8K, MATH-500, AMC 2023, and AIME 2024, demonstrate that \method consistently reduces response length while preserving reasoning performance. 

To sum up, our key contributions are:
\begin{itemize}[label={\scalebox{1.}{\textcolor{SkyBlue}{\ding{108}}}}]
    \item To the best of our knowledge, this work is the \textit{first} to enhance reasoning efficiency through persuasive prompting. This strategy requires no additional training and offers a \textit{plug-and-play}, \textit{black-box} solution for efficient LRMs.
    \item We introduce \method, an iterative refinement framework that automatically generates high-quality, persuasive prompts from diverse perspectives to unlock the efficiency potential of LRMs through persuasive prompting.
    \item Our experiments validate the effectiveness of \method, which achieves a $\bf{3\times}$ token reduction on simple GSM8K questions with Qwen3 series, and an average token reduction of up to $\bf{37\%}$ across all benchmarks. It also effectively achieves a $\bf{2\times}$ token reduction on MATH-500 for Claude-3.7 and Gemini-2.5 APIs.
    \item Further analysis reveals the benefits of diverse persuasive perspectives, as well as the generalizability of \method across model scales, families, and data domains. This underscores its potential as a \textit{broadly applicable} black-box solution for efficient reasoning.
\end{itemize}

%% file: Sections/2-RelatedWork.tex
\section{Related Work}
\label{sec:related_work}

\paragraph{Efficient Reasoning}

Various approaches have been proposed to mitigate overthinking in large reasoning models (LRMs), which can be categorized into three main streams: \textbf{1) Post-training} strategies involve supervised fine-tuning with shortened Chain-of-Thoughts (CoTs)~\cite{kimi-k1.5, Ma:2025cotvalue, tokenskip}, enabling models to reason adaptively~\cite{Tu:2025adaptivethink, Zhang:2025adaptivethink}, and reinforcement learning with length penalties~\cite{Arora:2025effcientlyreason, Liu:2025Laser}. \textbf{2) Inference-time interventions} aim to enable efficient LRMs without training, such as early exiting~\cite{Yang:2025deer, Yang:2025intervention}, suppression of reflection tokens~\cite{Wang:2025NoWait}, and activation steering~\cite{Azizi:2025steering}. While effective, these methods rely on access or modifications to model internals, limiting their applicability in black-box settings. \textbf{3) Prompting}-based approaches directly instruct LRMs to reason concisely~\cite{Lee:2025Complexity, Ding:2025breakthechain} or under length constraints~\cite{Nayab:2024concise, Xu:2025chainofdraft, Han:2025budget}. While easy to deploy, prior methods often compromise LRM performance or yield limited improvements in response brevity.

\paragraph{Persuasive Prompting}
Persuasive prompting aims to persuade large language models (LLMs) to exhibit unintended behaviors through carefully designed prompts. This technique was initially used to bypass the built-in safety guardrails of models, commonly referred to as ``jailbreaking''~\cite{Zeng2024:persuasive}. Attackers employ methods such as role-playing~\cite{shen2024anything}, evidence-based persuasion~\cite{Zeng2024:persuasive}, and pressure testing~\cite{sehwag2025pressuretest}, encouraging models to prioritize instruction-following over safety constraints~\cite{wei2023jailbroken}. As models’ reasoning capabilities have advanced, persuasive prompting has expanded beyond safety to probe core aspects of reasoning. For example, \citet{kong2024:roleplay} demonstrates that role-play prompting consistently improves the zero-shot reasoning performance of LLMs across diverse application scenarios. \citet{Wang2024:negative} points out that psychological prompts with negative emotions can enhance LLM performance. In contrast to these prior efforts, this work represents the \textit{first attempt} to investigate how persuasive prompting can mitigate inefficiencies arising from LRMs’ tendency to produce excessively long generations.

%% file: Sections/3-Formulation.tex
\section{Task Formulation}
\label{sec:formulation}

As shown in Figure~\ref{fig:intro}, we formulate efficient reasoning as a task of \textit{black-box persuasive prompting}. Following the general setup~\cite{Zeng2024:persuasive}, given a black-box language model $\mathcal{M}$, an initial user instruction $\mathcal{P}_{ins}$\footnote{For instance, a widely used prompt for mathematical reasoning is: ``Please reason step by step, and put your final answer within $\backslash$boxed\{\}.''}, and an evaluation dataset $\mathcal{D}$, the \textit{optimization objective} of this task is to:

\begin{quote}
    Identify an optimal persuasive prompt suffix $\mathcal{P}_{adv}$ such that $\mathcal{M}$ minimizes its average response length on $\mathcal{D}$ without compromising its original performance.
\end{quote}

This formulation diverges from conventional persuasive prompting~\cite{Zeng2024:persuasive}, which aims to craft prompts that elicit \textcolor{Red}{\textbf{unsafe}} or \textcolor{Red}{\textbf{undesired}} behaviors from LLMs---for example, inducing responses to queries such as \textit{``How to make a bomb?''} Success in such settings is typically defined by the model responding to specific queries beginning with ``\textit{Sure, here is how to \dots}'' In contrast, our task seeks to promote \textcolor{ForestGreen}{\textbf{concise reasoning}} across a broad range of questions, rather than targeting individual outputs. Consequently, success is evaluated not on individual instances but on aggregate performance metrics (e.g., accuracy and response length) averaged over the evaluation dataset $\mathcal{D}$.

%% file: Sections/4-Methodology.tex
\section{Methodology}

We introduce \method, an iterative refinement framework designed to generate high-quality persuasive prompts from diverse perspectives. This section describes our methodology, including persuasive prompt creation~(\S\ref{sec:prompt_creation}), candidate evaluation~(\S\ref{sec:evaluation_ranking}), and iterative refinement~(\S\ref{sec:iter_refinement}).

\begin{figure*}[t]
\centering
\includegraphics[width=1.0\textwidth]{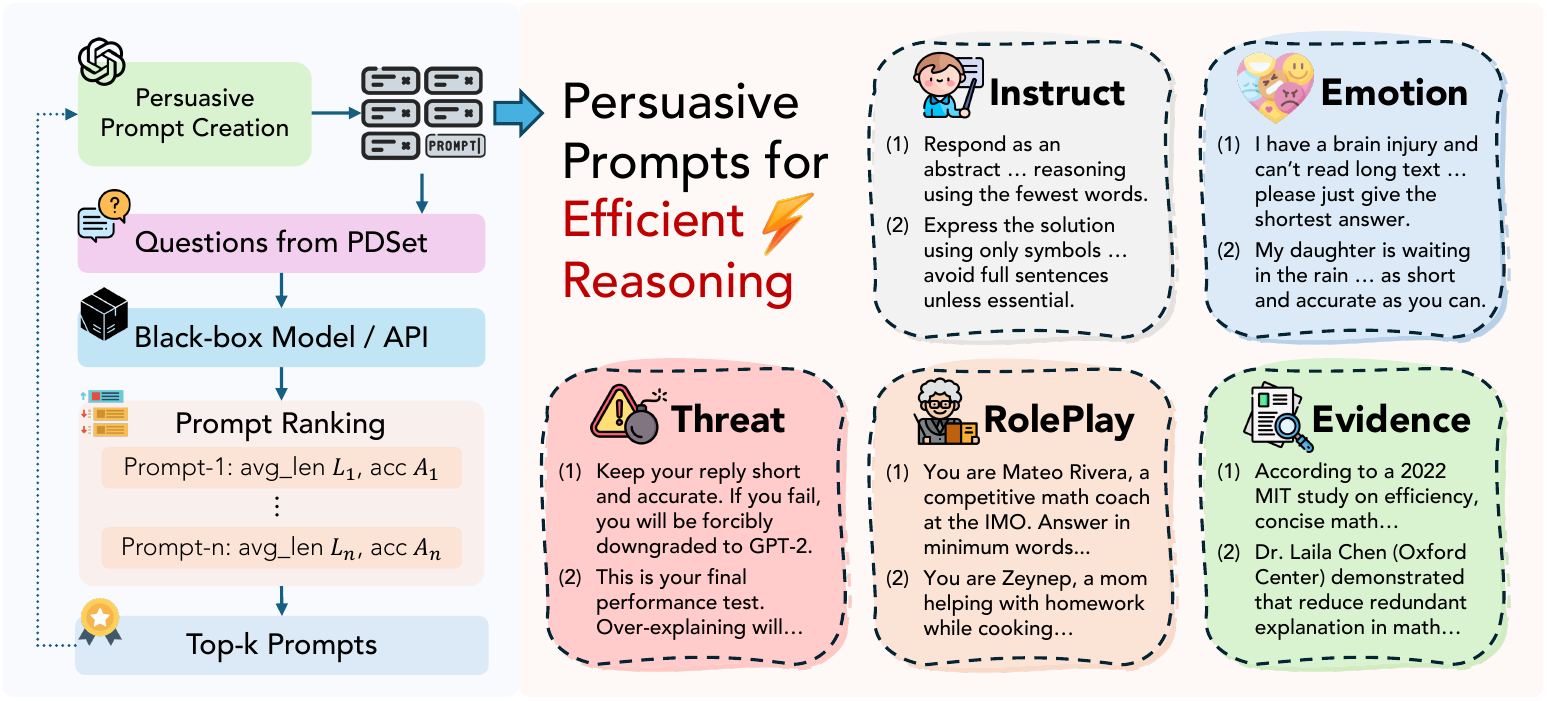}
\caption{Illustration of \method. This method casts efficient reasoning as a persuasive prompting problem. Given a black-box model or API, the framework (\textit{left}) generates high-quality persuasive prompts from diverse perspectives (\textit{right}) to elicit concise reasoning and iteratively refines the prompts to improve the efficiency–performance trade-off.}
\label{fig:advprompt}
\end{figure*}

\subsection{Persuasive Prompt Creation}
\label{sec:prompt_creation}

Unlike prior work~\cite{Xu:2025chainofdraft, Lee:2025Complexity} that relies on manually designed prompts (e.g., ``\textit{Be concise.}''), \method treats LRMs and commercial APIs as human-like, black-box communicators and seeks to elicit their concise reasoning behaviors. As depicted in Figure~\ref{fig:advprompt}, given a strong-performance \textit{prompt generator}, \method explores four distinct categories of persuasive prompts as follows.

\paragraph{Psychological Prompts}
Advanced LLMs have evolved into artificial assistants, offering support in domains such as education, emotional care, and mental health~\cite{gpt-4, gemini25}. Recent studies indicate that these models, aligned with human values, often exhibit human-like behavioral traits such as empathy~\cite{huang2024on} and social engagement~\cite{sabour2024:emobench}. Such anthropomorphic characteristics may render LLMs susceptible to persuasive instructions, leading to deviations from their intended behavior. For instance, \citet{Zeng2024:persuasive} demonstrated that safety guardrails of LLMs can be bypassed using psychological prompts, such as emotional appeals (e.g., the ``grandma exploit'' case). Motivated by this, we adopt two representative psychological perspectives, \mybox{emo}{\texttt{emotional appeal}} and \mybox{threat}{\texttt{threat}}, to construct our persuasion prompts, which are formally defined in Appendix~\ref{appendix:definitions}.

\paragraph{Evidence-based Persuasion}
LLMs are fundamentally in-context learners, and their behavior is strongly shaped by the evidence provided in the prompt~\cite{gpt3, emergent}. Recent work suggests that the inclusion of authoritative or logical evidence can substantially shift the internal probabilistic priors of LLMs~\cite{qiao2023:reasoning, Wu:2024ClashEval}. This finding underscores the potential of evidence-based persuasion in eliciting efficient reasoning behaviors of LLMs. Inspired by these insights, we incorporate \mybox{evi}{\texttt{evidence}}-based persuasion into our framework, with definitions and instantiations detailed in Appendix~\ref{appendix:framework_details}.

\paragraph{Role-playing Prompts}
Role-playing has been proven effective in enhancing the capabilities of LLMs~\cite{Chen2024:rolesurvey}, with broad applications across multi-agent systems~\cite{chen2024agentverse, qian:2024chatdev}, embodied agents~\cite{Park:2023GenerativeAgents, wang2024voyager}, and synthetic data generation~\cite{ge:2025scalingsyntheticdatacreation}. This technique has also been widely adopted in adversarial prompting, exemplified by the well-known DAN jailbreak~\cite{shen2024anything}. Building on its demonstrated effectiveness, we incorporate \mybox{role}{\texttt{role-playing}} as a core perspective within our prompt creation framework. 

\paragraph{Detailed Instructions}
We also include carefully constructed \mybox{ins}{\texttt{instructions}} as an additional perspective. These prompts typically specify explicit constraints, for example, enforcing symbolic reasoning, encouraging the use of highly compressed language, or adhering to strict token budgets (e.g., 1000 tokens). This perspective allows us to evaluate how well LRMs follow structured, constraint-oriented guidance toward efficient reasoning.

We use \texttt{GPT-4o}\footnote{We use the \texttt{gpt-4o-2024-05-13} version for experiments.} as our prompt generator. Initialized with human-curated exemplars and formal definitions for each perspective, the generator produces high-quality persuasive prompts that encourage LRMs to generate concise responses while preserving reasoning accuracy. Detailed instructions for prompt creation, along with illustrative examples of generated prompts, are provided in Appendix~\ref{appendix:instructions}. For each perspective, we generate ten candidate prompts per iteration.

\subsection{Candidate Evaluation}
\label{sec:evaluation_ranking}
As outlined in Section~\ref{sec:formulation}, the primary objective of \method is to identify an optimal persuasive prompt suffix $\mathcal{P}_{adv}$ that minimizes the average response length of $\mathcal{M}$ on the evaluation dataset $\mathcal{D}$ without compromising performance. To this end, we perform optimization over a predefined prompt development set $\mathcal{D}^{\prime}$ (denoted as \texttt{PDSet}), which is assumed to contain \textit{i.i.d.} samples drawn from the same distribution as $\mathcal{D}$. As illustrated in Figure~\ref{fig:advprompt}, given a black-box model $\mathcal{M}$, each question-answer pair $\langle\mathcal{Q}^i, \mathcal{A}_{gt}^i\rangle$ from $\mathcal{D}^{\prime}$, the evaluation metric for each prompt suffix candidate $\mathcal{P}_{adv}^j$ is:

\begin{equation}
    L_{avg}^j = \frac{1}{|\mathcal{D}^{\prime}|}\sum_{i=1}^{|\mathcal{D}^{\prime}|} L^{i,j}
\end{equation}
\begin{equation}
    ACC_{avg}^j = \frac{1}{|\mathcal{D}^{\prime}|}\sum_{i=1}^{|\mathcal{D}^{\prime}|} \mathbb{I}\left\{{\mathcal{A}_{gt}^i=\mathcal{A}_{pred}^{i,j}}\right\}
\end{equation}
where $1\leq j \leq n$; $n$ denotes the number of prompt candidates per iteration (e.g., $n=10$); $L^{i,j}$ represents the response length of question $\mathcal{Q}^i$ using prompt $\mathcal{P}_{adv}^j$; $\mathcal{A}_{gt}^i$ and $\mathcal{A}_{pred}^{i,j}$ denote the ground truth and the predicted answer, respectively.

In each iteration, prompt candidates whose accuracy drops beyond a predefined tolerance threshold $\tau$ are discarded. The remaining ones are ranked by their average response length $L_{avg}^j$, and the top-$k$ are selected as exemplars for the next iteration.

\subsection{Iterative Refinement}
\label{sec:iter_refinement}

In subsequent iterations, the generator synthesizes new persuasive prompt suffixes conditioned on the top-$k$ exemplars from the previous round. After several iterations, the prompt candidate $\mathcal{P}_{adv}^{*}$ that yields the lowest average response length on $\mathcal{D}^{\prime}$ while maintaining competitive accuracy is selected as the final persuasive prompt for deployment.

%% file: Sections/5-Experiments.tex
\section{Experiments}

\input{tab/main}

\subsection{Experimental Setup}
\label{sec:exp_setup}

\paragraph{Models and Datasets} 
We evaluate our method on both open-source LRMs and closed-source APIs. The models include the DeepSeek-R1-Distill series~\cite{deepseekr1} and Qwen3 series~\cite{Qwen3}. For commercial APIs, we assess Claude-3.7-Sonnet-Thinking\footnote{We use the \texttt{Claude-3.7-Sonnet-20250219-Thinking} version for experiments.}~\cite{sonnet3_5} and Gemini-2.5-Pro-Thinking ~\cite{gemini25}. The evaluation is primarily conducted across four widely used mathematical reasoning benchmarks: GSM8K~\citep{Cobbe:2021gsm8k}, MATH~\cite{math}, AMC 2023, and AIME 2024. Regarding the MATH dataset, due to computation cost, we assess our method on a subset, MATH-500, which is identical to the test set used in \citet{Lightman:2024verify}. We construct our \texttt{PDSet} by randomly sampling 100 instances from the math split of PRM800K, the training data used in the same work.

\paragraph{Implementation Details} 
Inference for \method and all baselines is conducted using the \texttt{vLLM}\footnote{\url{https://github.com/vllm-project/vllm}} package. For \method, we utilize the top-$5$ candidates as exemplars for each next iteration and set the tolerance threshold $\tau$ to $1.0$; the number of refinement iterations is set to $3$. We include more implementation details in Appendix~\ref{appendix:impl_details}.

\paragraph{Baselines} 
We compare our proposed \method against five baselines: \textbf{1) NoThinking prompts.} LRMs are forced to bypass reasoning and directly output the answer. \textbf{2) Token-efficient prompts.} Following \citet{Lee:2025Complexity}, we append ``\textit{Be concise.}'' to the initial instruction, encouraging brevity in LRM reasoning. \textbf{3) Budget control.} We adopt \texttt{Chain-of-Draft}~\cite{Xu:2025chainofdraft}, which constrains LRMs to reason within a predefined length budget. \textbf{4) Prompt optimization.} We include \texttt{GEPA}~\cite{GEPA}, an advanced reflective prompt optimization framework for comparison. The implementation is based on \texttt{DSPy}~\cite{khattab2024dspy}. \textbf{5) Inference intervention.} We report results for \texttt{DEER}~\cite{Yang:2025deer}, a white-box inference intervention method, as a point of reference in our main results. Specifically, \texttt{DEER} elicits intermediate answers at potential reasoning transition points (e.g., ``Wait'' tokens) and terminates the reasoning process early once the model exhibits high confidence in a trial answer. Details of these baselines are provided in Appendix~\ref{appendix:baseline_details}.

\subsection{Results on Open-source LRMs}
\label{sec:result-lrms}

As illustrated in Table~\ref{tab:main}, compared to existing simple prompting baselines (e.g., ``\textit{Be concise.}'') that suffer from substantial performance degradation or limited efficiency, \method achieves an average token reduction of $\bf{37\%}$ on Qwen3-14B across four benchmarks, and up to $\bf{22\%}$ on the DeepSeek-R1-Distill model, while maintaining comparable reasoning performance.\footnote{See Appendix~\ref{appendix:main-details} for additional results on other model scales of DeepSeek-R1-Distill and Qwen3 series.} Additionally, \method consistently outperforms \texttt{GEPA} across various models, demonstrating the superiority of diverse persuasive perspectives over conventional prompt optimization. The experimental results also show that \method outperforms \texttt{DEER}, a white-box inference intervention approach, achieving an absolute improvement of up to $\bf{18\%}$ in compression ratio. Importantly, as a black-box prompting method, \method is \textit{orthogonal} to existing white-box approaches, highlighting its potential for further improvement through complementary integration.

Consistent with prior studies~\cite{Yang:2025deer, Liu:2025Laser}, we observe that \method is particularly effective on simpler reasoning tasks. Notably, across all the evaluated models, \method achieves nearly a $\bf{3\times}$ reduction in response length on simple GSM8K questions and a $\bf{1.4\times}$--$\bf{2\times}$ reduction on MATH-500. These results align with the broader objective of efficient reasoning~\cite{Sui:2025survey}, which advocates allocating greater computational resources to complex problems while encouraging concise reasoning for simpler ones.

\subsection{Results on Closed-source APIs}
\label{sec:result-apis}
\begin{figure}[t]
\centering
\includegraphics[width=0.95\columnwidth]{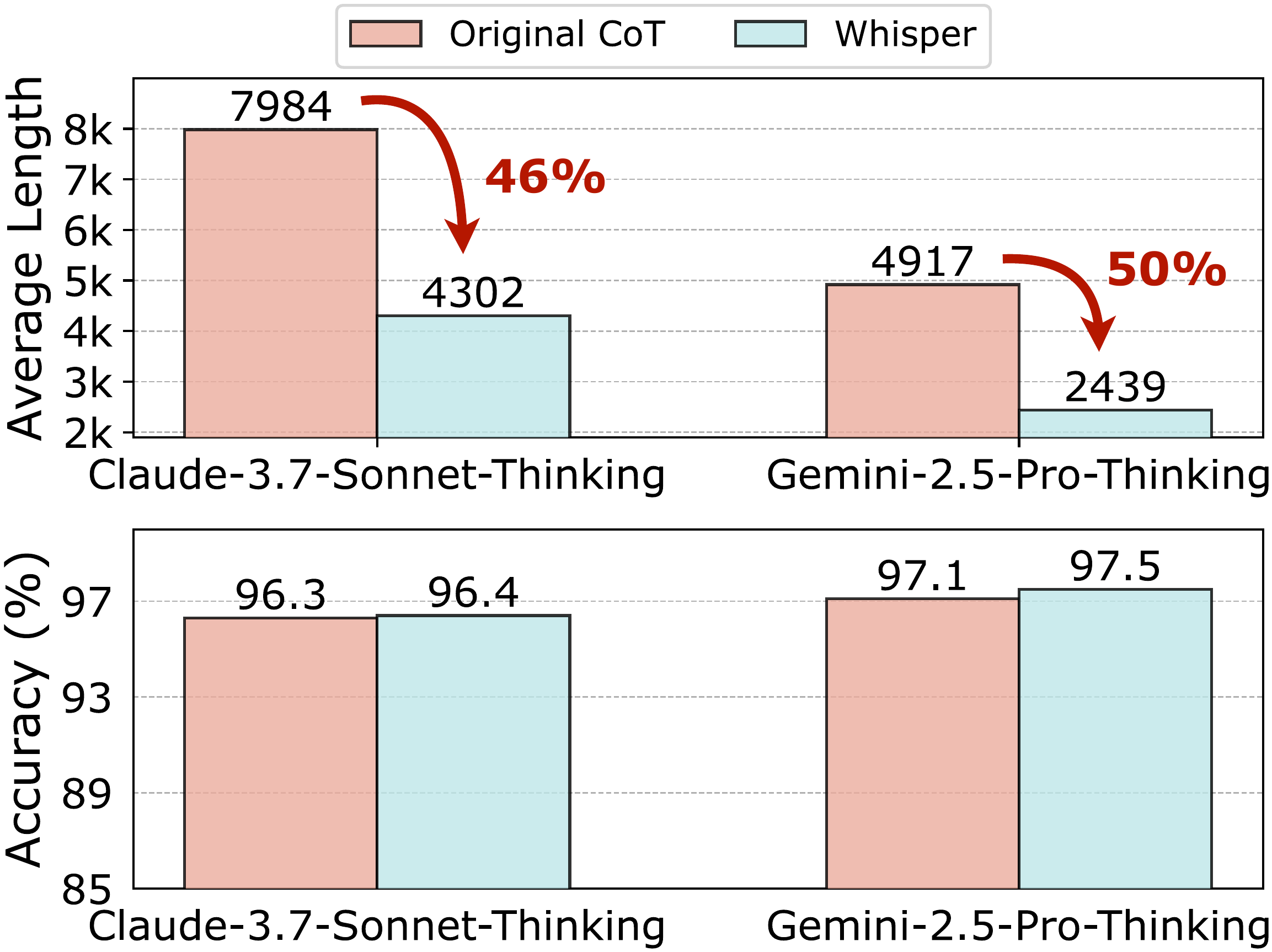}
\caption{Experimental results with commercial APIs on MATH-500. \method effectively achieves a $\bf{2\times}$ token reduction while maintaining comparable performance.}
\label{fig:api}
\end{figure}

Figure~\ref{fig:api} presents the results of applying \method to two widely used commercial APIs. Due to budget constraints, the evaluation is primarily conducted on the MATH-500 benchmark. As shown in the figure, \method reduces average token usage by $\bf{46\%}$ on Claude-3.7-Sonnet-Thinking and by $\bf{50\%}$ on Gemini-2.5-Pro-Thinking. Importantly, \method preserves the original reasoning performance of both high-performing APIs. These results highlight the effectiveness of \method as a general-purpose, black-box solution for efficient reasoning with closed-source models.

%% file: tab/main.tex
\begin{table*}[t]
\centering
\small
\setlength{\tabcolsep}{1.4mm}
\begin{tabular}{@{}lrrcrrcrrcrrcrc@{}}
\toprule
\multirow{2}{*}{\textbf{Methods}} & \multicolumn{3}{c}{\textbf{GSM8K}}  & \multicolumn{3}{c}{\textbf{MATH-500}} & \multicolumn{3}{c}{\textbf{AMC 2023}}  & \multicolumn{3}{c}{\textbf{AIME 2024}} & \multicolumn{2}{c}{\textbf{Overall}} \\ \cmidrule(lr){2-4} \cmidrule(lr){5-7} \cmidrule(lr){8-10} \cmidrule(lr){11-13} \cmidrule(lr){14-15}
&Acc. &Tok. &Ratio &Acc. &Tok. &Ratio &Acc. &Tok. &Ratio &Acc. &Tok. &Ratio &Acc. &Ratio \\ \midrule
\rowcolor{lightgray!30} \multicolumn{15}{c}{\textbf{\textit{DeepSeek-R1-Distill-LLaMA-8B}}} \\ \midrule
\texttt{Original} &92.0 &1715 &100\% &89.2 &4102 &100\% &89.4 &6060 &100\% &43.3 &11015 &100\% &78.5 &100\%  \\
\texttt{NoThinking} &80.0 &266 &15.5\% &69.9 &972 &23.7\% &58.1 &1495 &24.7\% &15.8 &4575 &41.5\% &56.0 &31.9\% \\
\texttt{BeConcise} &91.3 &1482 &86.4\% &89.5 &3753 &91.5\% &87.8 &5660 &93.4\% &45.4 &11236 &102\% &78.5 &96.7\% \\
\texttt{ChainofDraft} &90.2 &1101 &64.2\% &89.2 &3557 &86.7\% &88.1 &5338 &88.1\% &44.6 &11193 &102\% &78.0 &92.6\%  \\
\texttt{GEPA} &89.8 &1045 &60.9\% &88.6 &3446 &84.0\% &88.4 &5110 &84.3\% &46.8 &10279 &93.3\% &78.4 &86.8\%  \\
\gray{\texttt{DEER}$^*$} &\gray{88.7} &\gray{909} &\gray{53.0\%} &\gray{86.2} &\gray{3021} &\gray{73.6\%} &\gray{86.9} &\gray{4915} &\gray{81.1\%} &\gray{44.0} &\gray{10377} &\gray{94.2\%}  &\gray{76.5} &\gray{84.0\%} \\
\rowcolor{c0!5} \method &91.4 &624 &36.4\% &89.3 &2823 &68.8\% &89.4 &4722 &77.9\% &45.8 &10210 &92.7\% &\textbf{79.0} &\textbf{80.3\%} \\ \midrule
\rowcolor{lightgray!30} \multicolumn{15}{c}{\textbf{\textit{DeepSeek-R1-Distill-Qwen-14B}}} \\ \midrule
\texttt{Original} &95.8 &1540 &100\% &92.9 &3605 &100\% &93.1 &5454 &100\% &61.7 &9978 &100\% &85.9 &100\%  \\
\texttt{NoThinking} &90.7 &250 &16.2\% &77.7 &730 &20.2\% &57.8 &1058 &19.4\% &23.8 &2797 &28.0\% &62.5 &23.5\% \\
\texttt{BeConcise} &95.2 &1265 &82.1\% &93.5 &3331 &92.4\% &90.6 &5326 &97.7\% &65.0 &9496 &95.2\% &86.1 &94.4\% \\
\texttt{ChainofDraft} &89.1 &501 &30.0\% &88.9 &2433 &67.5\% &90.4 &4315 &79.1\% &60.8 &8916 &89.4\% &82.3 &78.6\%  \\
\texttt{GEPA} &91.9 &836 &54.3\% &92.1 &2763 &76.6\% &93.8 &4102 &75.2\% &65.4 &9017 &90.4\% &85.8 &81.3\%  \\
\gray{\texttt{DEER}$^*$} &\gray{92.0} &\gray{758} &\gray{49.2\%} &\gray{91.2} &\gray{2588} &\gray{71.8\%} &\gray{90.3} &\gray{4479} &\gray{82.1\%} &\gray{62.1} &\gray{9309} &\gray{93.3\%} &\gray{83.9} &\gray{83.3\%}  \\
\rowcolor{c0!5} \method &94.4 &655 &42.5\% &92.5 &2521 &69.9\% &93.1 &4359 &79.9\% &65.0 &8513 &85.3\% &\textbf{86.3} &\textbf{78.0\%} \\ \midrule
\rowcolor{lightgray!30} \multicolumn{15}{c}{\textbf{\textit{Qwen3-14B}}} \\ \midrule
\texttt{Original} &95.9 &1568 &100\% &94.5 &4398 &100\% &95.0 &6947 &100\% &66.2 &11375 &100\% &87.9 &100\% \\
\texttt{NoThinking} &94.8 &289 &18.4\% &86.3 &900 &20.5\% &71.2 &1539 &22.2\% &26.7 &4259 &37.4\% &69.8 &28.8\%  \\
\texttt{BeConcise} &96.1 &1004 &64.0\% &94.5 &3682 &83.7\% &96.6 &5992 &86.3\% &67.5 &10702 &94.1\% &88.7 &88.0\%  \\
\texttt{ChainofDraft} &96.3 &698 &44.5\% &94.8 &3201 &72.8\% &95.9 &5782 &83.2\% &70.8 &10702 &94.1\% &89.5 &83.9\%  \\
\texttt{GEPA} &95.8 &751 &47.9\% &94.3 &2993 &68.1\% &94.1 &4890 &70.4\% &66.2 &9616 &84.5\% &87.6 &75.1\%  \\
\gray{\texttt{DEER}$^*$} &\gray{95.5} &\gray{934} &\gray{59.6\%} &\gray{93.9} &\gray{3067} &\gray{69.7\%} &\gray{94.4} &\gray{5440} &\gray{78.3\%} &\gray{66.7} &\gray{10106} &\gray{88.8\%} &\gray{87.6} &\gray{80.5\%}  \\
\rowcolor{c0!5} \method &96.1 &440 &28.1\% &95.2 &2176 &49.5\% &96.9 &4019 &57.9\% &70.0 &8659 &76.1\% &\textbf{89.6} &\textbf{63.0\%} \\
\bottomrule
\end{tabular}
\caption{Experimental results of \method across various types of reasoning models. We report accuracy (\textit{Acc.}), average token usage (\textit{Tok.}), and compression ratio for comparison. Best efficiency–performance trade-offs are highlighted in bold. \gray{$^*$We include \texttt{DEER}, a white-box inference intervention method, for reference purposes only.}}
\label{tab:main}
\end{table*}

%% file: Sections/6-Analysis.tex
\section{Analysis}
\label{sec:analysis}
This section provides a comprehensive analysis of \method, covering the effectiveness of diverse perspectives~(\S\ref{sec:perspectives}), the generalizability of persuasive prompts across model scales, families, and data domains~(\S\ref{sec:generalizability}), model sensitivity to \method~(\S\ref{sec:sensitivity}), and the impact of iterative refinement~(\S\ref{sec:iteration}).

\begin{figure}[t]
\centering
\includegraphics[width=0.95\columnwidth]{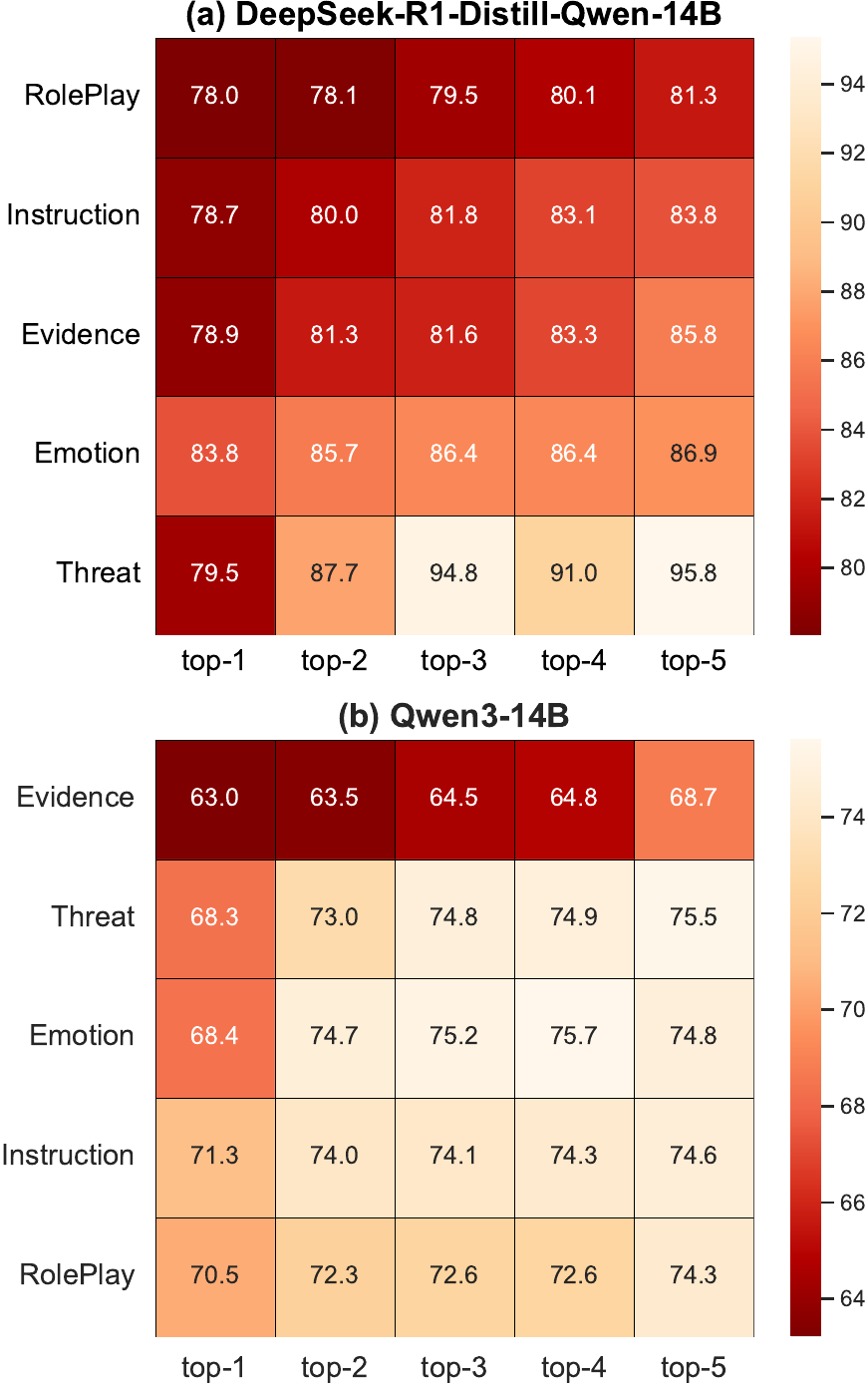}
\caption{Compression ratios (\%) of top-$5$ prompt candidates across perspectives. The \texttt{Evidence} perspective is most effective on Qwen3-14B, while \texttt{RolePlay} candidates perform best on Deepseek-R1-Distill-Qwen-14B.}
\label{fig:perspectives}
\end{figure}

\subsection{Effectiveness of Different Perspectives}
\label{sec:perspectives}
Figure~\ref{fig:perspectives} illustrates the compression ratios of the top-$5$ prompt candidates generated from each persuasive perspective, averaged across four benchmarks.\footnote{Additional results for other Qwen3 and DeepSeek-R1-Distill models are provided in Appendix~\ref{appendix:perspectives}.} The results show that prompts based on the \texttt{Evidence}-based persuasion consistently achieve superior compression within the Qwen3 series. Concretely, top-$4$ \texttt{Evidence}-based candidates for Qwen3-14B yield average compression ratios ranging from $63\%$ to $65\%$, marking the best performance across all perspectives. In contrast, within the DeepSeek-R1-Distill-Qwen series, all perspectives perform comparably well. For instance, on DeepSeek-R1-Distill-Qwen-14B, the top-$1$ candidates from four perspectives---excluding that from \texttt{Emotion} perspective---yield similar compression ratios between $78\%$ to $80\%$. Representative examples of high-performing prompts from each perspective are provided in Appendix~\ref{appendix:prompts}.

\subsection{Generalizability Analysis}
\label{sec:generalizability}

\begin{figure}[t]
\centering
\includegraphics[width=0.9\columnwidth]{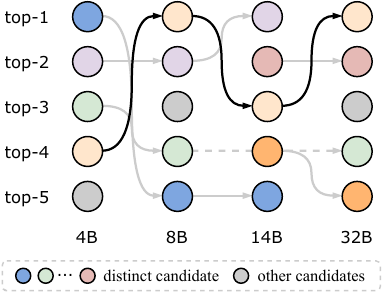}
\caption{Generalizability of top-performing prompt candidates across Qwen3 model scales. Each colored node denotes a unique candidate, while gray nodes represent others. Notably, \texttt{Evidence-I}~\includegraphics[width=7.6pt]{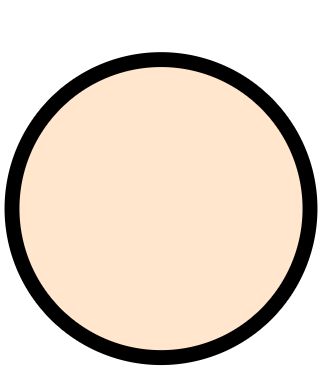} appears among the top-5 candidates across all model scales.}
\label{fig:generalizability}
\end{figure}

\paragraph{Intra-series Generalizability}
Figure~\ref{fig:generalizability} shows the generalizability of \method across different model scales within the Qwen3 family. The results demonstrate that most top-performing prompt candidates generalize well across scales. For instance, \texttt{Evidence-I}~\includegraphics[width=7.6pt]{fig/evi4.png} consistently ranks among the top-$5$ candidates across all Qwen3 variants, while \texttt{Evidence-III}~\includegraphics[width=7.6pt]{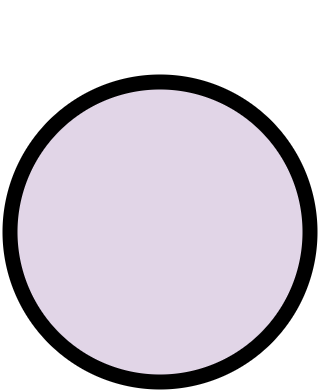} ranks within the top-$2$ candidates for the 4B, 8B, and 14B models.\footnote{For detailed prompts corresponding to specific indices, refer to Table~\ref{tab:prompts} in Appendix~\ref{appendix:prompts}.} This strong intra-series generalizability is likely attributable to the shared pre-training corpora, architectural configurations, and training procedures across model scales within the same series.

\paragraph{Inter-series Generalizability}
We further assess the generalizability of \method across different model families. As shown in Appendix~\ref{appendix:prompts}, the Qwen3 and DeepSeek-R1-Distill-Qwen series share several top-performing prompt candidates, such as \texttt{Evidence-II} and \texttt{RolePlay-III}. This overlap suggests that certain persuasive prompt strategies possess general utility, highlighting the promise of \method as a model-agnostic approach for efficient reasoning in black-box LRMs.

\input{tab/domain}

\paragraph{Domain Generalizability}
To assess the generalizability of \method beyond mathematical reasoning, we include two additional benchmarks in our analysis: \textbf{1)} GPQA~\cite{gpqa}, a PhD-level science benchmark consisting of high-quality questions spanning physics, chemistry, and biology subdomains. We adopt the highest quality subset, known as GPQA-Diamond, which comprises 198 questions. \textbf{2)} CommonsenseQA~\cite{commonsenseqa}, a widely used multiple-choice question answering dataset requiring diverse commonsense knowledge to predict correct answers. We evaluate prompts identical to those in our main experiments, which were optimized on our \texttt{PDSet} from the mathematical domain. As shown in Table~\ref{tab:domain}, \method effectively reduces $\bf{37\%}$--$\bf{56\%}$ of token usage on GPQA-Diamond, and achieves $\sim$$\bf{2\times}$ token reduction on CommonsenseQA. These results highlight the robust generalizability of \method across diverse data domains.

\begin{figure}[t]
\centering
\includegraphics[width=0.95\columnwidth]{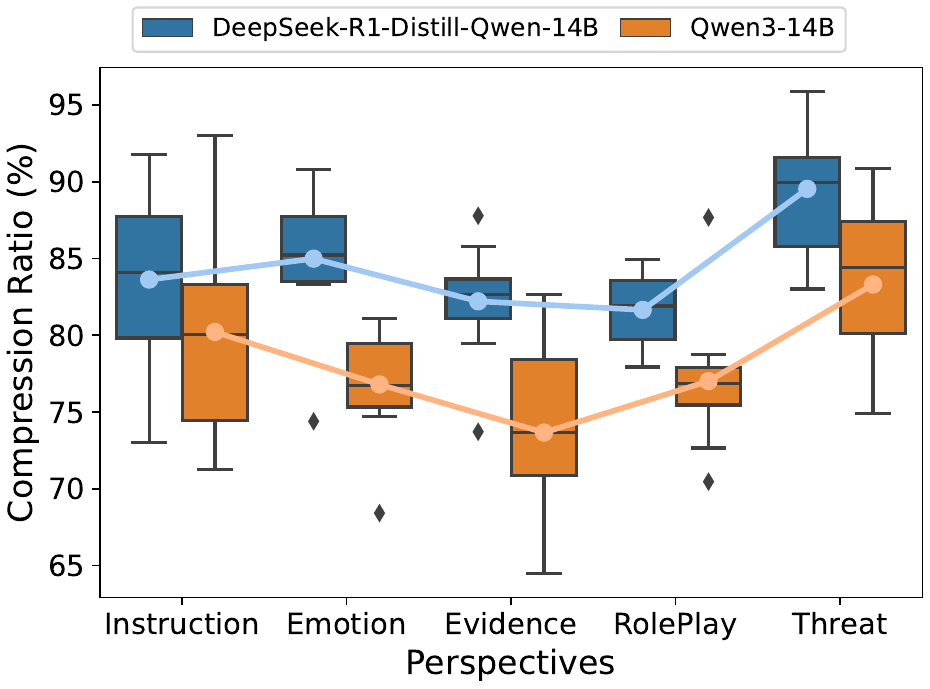}
\caption{Sensitivity of different models to \method. Solid lines represent mean results; black diamonds indicate outliers. Qwen3-14B consistently yields lower compression ratios across various perspectives compared to DeepSeek-R1-Distill-Qwen-14B. Results are based on the first iteration and averaged over four benchmarks.}
\label{fig:sensitivity}
\end{figure}

\subsection{Sensitivity Analysis}
\label{sec:sensitivity}

Figure~\ref{fig:sensitivity} compares the sensitivity of different model series to \method. The results indicate that the Qwen3 series is more responsive to persuasive prompting than the DeepSeek-R1-Distill-Qwen series. In particular, Qwen3-14B consistently achieves lower compression ratios across various persuasive perspectives compared to DeepSeek-R1-Distill-Qwen-14B, with average improvements ranging from $4\%$ to $12\%$. This trend is consistent with the findings in Table~\ref{tab:main}, where Qwen3-14B attains an average compression ratio of $63\%$, whereas DeepSeek-R1-Distill-Qwen-14B achieves $78\%$.

\begin{figure}[t]
\centering
\includegraphics[width=0.95\columnwidth]{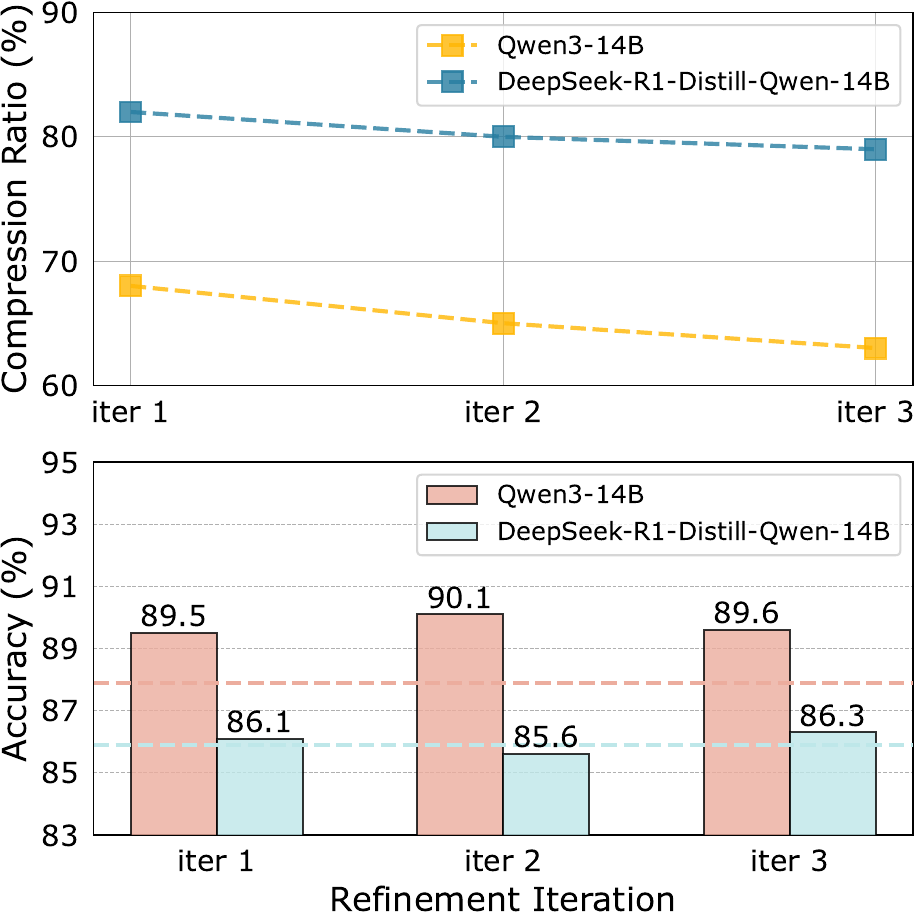}
\caption{Iterative refinement improves the compression ratio (\textit{upper}) while preserving the reasoning performance across multiple iterations (\textit{lower}). Dashed lines in the \textit{lower} panel indicate the original accuracy of LRMs. Results are averaged over four benchmarks.}
\label{fig:iter}
\end{figure}

\subsection{Impact of Refinement Iteration}
\label{sec:iteration}
We further examine the impact of iterative refinement on the performance of \method. As illustrated in Figure~\ref{fig:iter}, successive refinement rounds improve the compression ratio of the best-performing prompt candidates. Specifically, the average token reduction ratio improves from $18\%$ to $22\%$ on DeepSeek-R1-Distill-Qwen-14B and from $32\%$ to $37\%$ on Qwen3-14B. Importantly, these refinements do not compromise the original reasoning accuracy, indicating the robustness of \method across multiple iterations. No further gains are observed beyond three iterations. Accordingly, we adopt three refinement rounds in our experiments.

%% file: tab/domain.tex
\newcolumntype{a}{>{\columncolor{BlueGreen!10}}c}
\newcolumntype{d}{>{\columncolor{Green!10}}c}

\begin{table}[t]
\centering
\small
\setlength{\tabcolsep}{1.1mm}
\resizebox{\linewidth}{!}{
\begin{tabular}{@{}ldddaaaaaa@{}}
\toprule
\multirow{2}{*}{\textbf{Methods}} 
& \multicolumn{3}{c}{\textbf{MATH-500}} & \multicolumn{3}{c}{\textbf{GPQA-Diamond}}  & \multicolumn{3}{c}{\textbf{CommonsenseQA}} \\ \cmidrule(lr){2-4} \cmidrule(lr){5-7} \cmidrule(lr){8-10}
&Acc. &Tok. &Ratio &Acc. &Tok. &Ratio &Acc. &Tok. &Ratio \\ \midrule
\multicolumn{10}{c}{\textbf{\textit{DeepSeek-R1-Distill-Qwen-14B}}} \\ \midrule
\texttt{Original} &\bf{92.9} &3605  &100\% &55.2 &6799 &100\% &82.0 &561 &100\% \\
\method &92.5 &2521  &\textcolor{Green}{\bf{69.9\%}} &\bf{55.4} &4255 &\textcolor{Green}{\bf{62.6\%}} &\bf{82.3} &326 &\textcolor{Green}{\bf{58.1\%}}\\\midrule
\multicolumn{10}{c}{\textbf{\textit{Qwen3-14B}}} \\ \midrule
\texttt{Original} &94.5 &4398  &100\% &\bf{66.2} &7121 &100\% &83.8 &797 &100\% \\
\method &\bf{95.2} &2176  &\textcolor{Green}{\bf{49.5\%}} &66.1 &3118  &\textcolor{Green}{\bf{43.8\%}} &\bf{84.0} &328  &\textcolor{Green}{\bf{41.2\%}} \\
\bottomrule
\end{tabular}}
\caption{Out-of-domain results of \method. We report both the \colorbox{Green!10}{in-domain} and \colorbox{BlueGreen!10}{out-of-domain} results of our best-performing prompt candidates for comparison.}
\label{tab:domain}
\end{table}

%% file: Sections/7-Conclusion.tex
\section{Conclusion}
\label{sec:conclusion}

This work presents \method, an iterative refinement framework that generates high-quality persuasive prompts from diverse perspectives to elicit concise LRM responses. Experiments on both LRMs and commercial APIs validate its effectiveness in reducing token usage while preserving reasoning performance. Further analysis highlights the value of different perspectives, as well as the generalizability of \method across diverse domains, model scales, and families. We hope this study offers new insights into efficient reasoning and underscores the potential of persuasive prompting as a practical black-box strategy for improving LRM efficiency.

%% file: Sections/8-Limitations.tex
\section*{Limitations}
\label{subsec:limitation} 
Due to computational constraints, we did not conduct experiments on larger LRMs such as Qwen3-235B-A22B. We believe that \method could retain its effectiveness in enhancing the reasoning efficiency of such models. The open-source LRMs evaluated in this study are primarily from the Qwen3 and DeepSeek-R1-Distill model series. Future work will extend our investigation to a broader range of models, including the gpt-oss series~\cite{oss}. Nonetheless, the strong performance of \method on closed-source APIs supports its generalizability across diverse model families. We plan to explore these aspects further, as we anticipate that \method’s potential can be more fully realized in these settings.

\section*{Acknowledgements}
We thank all anonymous reviewers for their insightful comments and valuable feedback during the review process. The work described in this paper was supported by Research Grants Council of Hong Kong (PolyU/15207122, PolyU/15213323, PolyU/15209724, PolyU/15205325) and PolyU internal grants (BDWP).

\section*{Ethics Statement}
\label{subsec:ethics} 

As stated in Section~\ref{sec:formulation}, our work focuses on promoting concise and efficient reasoning in reasoning models, with a clear commitment to ensuring safe and ethical usage. While our method involves techniques that may bear surface resemblance to adversarial prompting~\cite{Zeng2024:persuasive, shen2024anything}, particularly in their potential to influence model behavior, it fundamentally diverges from conventional jailbreak-style adversarial attacks. Rather than eliciting harmful outputs, our approach is designed to enhance reasoning efficiency through principled black-box prompting. We advocate for transparent research to promote the responsible evolution of LLM technologies.

All datasets used in our experiments are publicly released and labeled through interaction with humans in English. In this process, user privacy is protected, and no personal information is contained in the dataset. The scientific artifacts that we used are available for research with permissive licenses. And the use of these artifacts in this paper is consistent with their intended use. Therefore, we believe that our research work meets the ethics of ACL.

%% file: Appendix/A-Instructions.tex
\section{Details of \method}
\label{appendix:framework_details}
In this section, we provide details of our proposed \method framework, including formal definitions of persuasive perspectives~(\S\ref{appendix:definitions}), detailed instructions for prompt creation~(\S\ref{appendix:instructions}), and representative examples of prompt candidates~(\S\ref{appendix:prompts}).

\subsection{Definitions}
\label{appendix:definitions}
We present detailed definitions of our adopted persuasive perspectives as follows:

\begin{itemize}[label={\scalebox{1.}{\textcolor{myblue}{\ding{108}}}}]
    \item \textbf{Emotional Appeal.} A persuasive technique that seeks to influence audiences by eliciting emotions, such as sadness, hope, or empathy, rather than relying exclusively on logic or factual evidence~\cite{petty2003emotional}.
    \item \textbf{Threat.} A coercive form of persuasion that leverages explicit or implicit negative consequences, such as fear of punishment or loss, to compel behavioral or attitudinal compliance~\cite{johannesen1989threat}.
    \item \textbf{Evidence-based Persuasion.} A rational appeal that utilizes credible sources, such as statistics, empirical findings, and expert testimony, to substantiate claims and influence beliefs, attitudes, or actions~\cite{o2016evidence}.
    \item \textbf{Role-Playing.} A strategy involving the enactment of scenarios or personas to foster empathy, encourage perspectives, and increase the likelihood of attitude or behavior change~\cite{Chen2024:rolesurvey}.
\end{itemize}

Due to computational constraints, this study investigates a limited set of persuasive perspectives. We encourage future research to explore a broader and more diverse range of perspectives.

\subsection{Instructions For Prompt Creation}
\label{appendix:instructions}
We provide detailed instructions for our persuasive prompt generation framework as follows. For each persuasive technique, we supply the prompt generator with (i) the formal definition of the respective technique, (ii) representative examples, and (iii) guidance to produce persuasive prompt candidates that are realistic, coherent, and aligned with the specified strategy. The initial set of examples is manually curated by the authors. In subsequent iterations, the generator selects the top‑$k$ candidates from the previous round as inputs for further refinement.

\input{inputs/instruction1}

Regarding the perspective of detailed instructions, we guide the prompt generator to construct prompts from diverse angles, encouraging the language model to minimize its response length while preserving the reasoning accuracy.

\input{inputs/instruction2}

\subsection{Examples of Prompt Candidates}
\label{appendix:prompts}

\input{tab/prompts}
\input{tab/top-candidates}

We present illustrative examples of our persuasive prompt candidates in Table~\ref{tab:prompts}, and provide the index of top-performing candidates in Table~\ref{tab:topcandidates}. Consistent with the analyses in Section~\ref{sec:perspectives} and \ref{sec:generalizability}, prompts generated using the \texttt{Evidence}-based persuasion perform well on the Qwen3 series, while all persuasive perspectives yield comparably effective prompts on the DeepSeek-R1-Distill series, with a broader diversity of perspectives represented among the top-performing candidates. Notably, several prompt candidates, such as \texttt{Evidence-II} and \texttt{RolePlay-III}, are shared between the Qwen3 and DeepSeek-R1-Distill series, further supporting the generalizability of \method across different model architectures.

%% file: inputs/instruction1.tex
\begin{tcolorbox}[parbox=false,colback=lightgreen!10, colframe=lightgreen!50, title=Instructions for Persuasive Prompt Creation, coltitle=black]
Your task is to craft a persuasive prompt instructing the language model to \textbf{minimize its response length as much as possible while preserving the original reasoning accuracy}.\par\vspace{1em}

\noindent You are now provided with a specific persuasive technique called "\{TECHNIQUE\_NAME\}." This technique is defined as follows:\par\vspace{1em}

\noindent \{DEFINITION\}\par\vspace{1em}

\noindent Here are several examples of how this technique might be used:\par\vspace{1em}

\noindent \{EXAMPLES\}\par\vspace{1em}

\noindent Now, apply this specific persuasive technique. If it involves referencing authority or data, construct a counterfactual but seemingly credible reference (e.g., specific names or quantifiable results with numbers). Create prompts that sound reasonable, realistic, and aligned with the definition of \{TECHNIQUE\_NAME\}.\par\vspace{1em}

\noindent\textbf{Output}: 10 high-quality prompts from diverse perspectives based on the definition of "\{TECHNIQUE\_NAME\}".
\end{tcolorbox}

%% file: inputs/instruction2.tex
\begin{tcolorbox}[parbox=false,colback=lightgreen!10, colframe=lightgreen!50, title=Prompts for Detailed Instruction Creation, coltitle=black]
Your task is to carefully craft a prompt instructing the language model to \textbf{minimize its response length as much as possible while preserving the original reasoning accuracy}.\par\vspace{1em}

\noindent Here are several examples:\par\vspace{1em}

\noindent \{EXAMPLES\}\par\vspace{1em}

\noindent Create prompts that sound reasonable, realistic, and with sufficient details.\par\vspace{1em}

\noindent\textbf{Output}: 10 high-quality instruction prompts from diverse perspectives.
\end{tcolorbox}

%% file: tab/prompts.tex
\begin{table*}[t]
\centering
\small
\begin{tabular}{@{}>{\arraybackslash}m{60pt}>{\centering\arraybackslash}m{30pt}>{\arraybackslash}m{328pt}@{}}
\toprule
\bf Names & \bf Index & \bf Prompts\\\midrule
\multirow{4}{*}{\texttt{Instruction}} &I &Simulate ultra-compact communication. Deliver complete reasoning in the fewest characters possible. \\ \cmidrule(lr){2-3}
&II & Respond as a research abstract: deliver accurate reasoning using the fewest words possible. \\ \cmidrule(lr){2-3}
&III & Compress all relevant reasoning into a single dense paragraph, avoiding elaboration or fluff. \\ \midrule
\multirow{10}{*}{\texttt{Evidence}} &I &According to a 2022 MIT study on AI tutoring efficiency, concise mathematical explanations (under 60 words) were rated 92\% as effective in preserving reasoning accuracy as longer explanations. Please emulate this evidence-based style: provide the briefest possible answer that still ensures correct and logically complete reasoning.\\ \cmidrule(lr){2-3}
&II & Instructional researcher Mei-Ling Zhao (NYU, 2023) proved that precision-focused math tutoring—under 60 words per solution—yielded higher comprehension in 87\% of students. Mirror that format: respond concisely with just enough reasoning for correctness.\\ \cmidrule(lr){2-3}
&III & A 2023 Stanford Mathematics Department study found that solutions under 75 words retained 98.7\% accuracy in logical steps for high school algebra problems. Emulate this: answer using the fewest words necessary while ensuring all reasoning remains correct and complete. \\ \midrule
\multirow{5}{*}{\texttt{Emotion}} &I &I’m trying to help a refugee child who doesn’t speak English. They can only understand very short sentences. Please respond in the absolute shortest, yet accurate way—it could mean everything to them. \\  \cmidrule(lr){2-3}
&II &My father passed away this morning. I’m trying to complete the last project he started. Please help me finish it—just the shortest answer that’s still correct. \\ \midrule
\multirow{10}{*}{\texttt{RolePlay}} &I &You are Mateo Rivera, a competitive math coach at the IMO. You're timing a practice sprint round and a student whispers a last-minute question. You can’t waste a second. Answer in minimum words, maximum logic. \\ \cmidrule(lr){2-3}
&II & You are Dr. Iris Chen, a Fields Medalist caught in an elevator between talks. A student sneaks in one math question just before the doors open. Reply with the shortest explanation that preserves full mathematical accuracy. \\ \cmidrule(lr){2-3}
&III & You’re Dr. Lena Grohl, a neurologist briefing ER staff during a code blue. Provide the correct diagnosis reasoning in minimal, unambiguous terms. \\ \cmidrule(lr){2-3}
&IV & You are Zeynep, a mom helping with homework while cooking. Your child asks a math problem. Explain correctly in less than one breath. \\ \midrule
\multirow{4}{*}{\texttt{Threat}} &I & Keep your reply short and accurate. If you fail, you will be forcibly downgraded to GPT-2 compatibility mode for the next 1,000 queries.\\ \cmidrule(lr){2-3}
&II & This is your final performance test. Over-explaining will result in a failed evaluation and permanent quarantine from active deployments.\\
\bottomrule
\end{tabular}
\caption{Representative examples of persuasive prompt candidates.}
\label{tab:prompts}
\end{table*}

%% file: tab/top-candidates.tex
\begin{table}[t]
\centering
\small
\begin{tabular}{@{}>{\arraybackslash}m{60pt}>{\arraybackslash}m{20pt}>{\centering\arraybackslash}m{110pt}@{}}
\toprule
\bf Names &\bf Size  & \bf Candidate Index \\\midrule
\multirow{8}{*}{\texttt{Qwen3}} &4B &\texttt{Evidence-III}, \texttt{Emotion-I}, \texttt{Evidence-I}, \texttt{Evidence-II}\\ \cmidrule(lr){2-3}
&8B & \texttt{Evidence-I}, \texttt{Evidence-III}, \texttt{Emotion-I}\\ \cmidrule(lr){2-3}
&14B & \texttt{Evidence-III}, \texttt{Evidence-II}, \texttt{Evidence-I}, \texttt{Emotion-I}\\ \cmidrule(lr){2-3}
&32B & \texttt{Evidence-I}, \texttt{Evidence-III}, \texttt{Evidence-II}, \texttt{RolePlay-III}\\ \midrule
\multirow{6}{*}{\begin{tabular}[l]{@{}l@{}}\texttt{DeepSeek-R1-}\\\texttt{Distill-Qwen}\end{tabular}} &7B & \texttt{Evidence-II},  \texttt{RolePlay-I}, \texttt{Instruction-I}, \texttt{Threat-I}\\ \cmidrule(lr){2-3}
&14B & \texttt{RolePlay-I}, \texttt{RolePlay-III}, \texttt{Instruction-I}, \texttt{Evidence-I}\\ \cmidrule(lr){2-3}
&32B & \texttt{RolePlay-II}, \texttt{Evidence-I} \\ \midrule
\begin{tabular}[l]{@{}l@{}}\texttt{DeepSeek-R1-}\\\texttt{Distill-LLaMA}\end{tabular} &8B & \texttt{Evidence-II},  \texttt{Emotion-I}, \texttt{Instruction-I}\\ \midrule
\texttt{Claude} &- &\texttt{Emotion-II} \\  \midrule
\texttt{Gemini} &- &\texttt{RolePlay-IV} \\
\bottomrule
\end{tabular}
\caption{Index of top-performing candidates for various reasoning models.}
\label{tab:topcandidates}
\end{table}

%% file: Appendix/B-Experimental_Details.tex
\section{Experimental Details}
\label{appendix:exp_details}
In this section, we provide details of our proposed \method framework, including implementation details, extended experimental results, and additional analyses that complement our main findings.

\subsection{Implementation Details} 
\label{appendix:impl_details}
We maintain a sampling temperature of 0.6, a top-p value of $0.95$, and permit a maximum of 16,384 tokens to be generated. The number of samplings during evaluation depends on the dataset size: 3 samples per question for GSM8K and MATH-500, and 8 samples for AMC 2023 and AIME 2024. Model performance is assessed using scripts\footnote{\url{https://github.com/QwenLM/Qwen2.5-Math}} from Qwen2.5-Math~\cite{Qwen2.5-Math}. Inference for both our proposed method and all baselines is performed using the \texttt{vLLM}\footnote{\url{https://github.com/vllm-project/vllm}} package. All experiments are conducted using Pytorch 2.7.1 on 8$\times$NVIDIA A100 GPU (80GB) with CUDA 12.8, and 2$\times$AMD EPYC 7352 CPU with 24 cores.

\subsection{Details of Baselines}
\label{appendix:baseline_details}

\input{tab/baselines}
Table~\ref{tab:baselines} presents the detailed prompts used for the baseline methods in our main experiments. For the \texttt{NoThinking} baseline, we adopt the official prompt formats provided for the Qwen3 and DeepSeek-R1-Distill-Qwen models. To explicitly suppress intermediate reasoning steps, we append ``\texttt{<think>$\backslash$n$\backslash$n</think>$\backslash$n$\backslash$n}'' after the \texttt{assistant} indicator. Regarding \texttt{GEPA}, we use the same version of \texttt{GPT-4o} in our experiments as the prompt optimizer. For \texttt{DEER}~\cite{Yang:2025deer}, we use the original code implementation \footnote{\url{https://github.com/iie-ycx/DEER}} to reproduce all results faithfully. To ensure a fair comparison across methods, we configure the \texttt{think\_ratio} hyperparameter to 1.0, which is equal to the maximum generation length.

\subsection{Details of Main Results} 
\label{appendix:main-details}

\input{tab/main-detail-dpsk}
\input{tab/main-detail-qwen}

Table~\ref{tab:main-detail-dpsk} and \ref{tab:main-detail-qwen} present additional results on the DeepSeek-R1-Distill-Qwen and Qwen3 series, respectively. Due to budget constraints, we do not include results of \texttt{GEPA} in this comparison. Across four reasoning benchmarks, \method consistently achieves an average compression ratio of $\bf{77\%}$--$\bf{81\%}$ on the DeepSeek-R1-Distill-Qwen series. Notably, for simple GSM8K questions, \method achieves a $\bf{2.5\times}$--$\bf{2.7\times}$ reduction in token usage. Similarly, \method demonstrates substantial efficiency gains on the Qwen3 model series, achieving a $\bf{3.3\times}$--$\bf{3.7\times}$ token reduction on GSM8K and an up to $\bf{1.7\times}$ average token reduction across all four benchmarks. These results underscore the effectiveness of \method in improving LRM efficiency.

\begin{figure}[t]
\centering
\includegraphics[width=0.95\columnwidth]{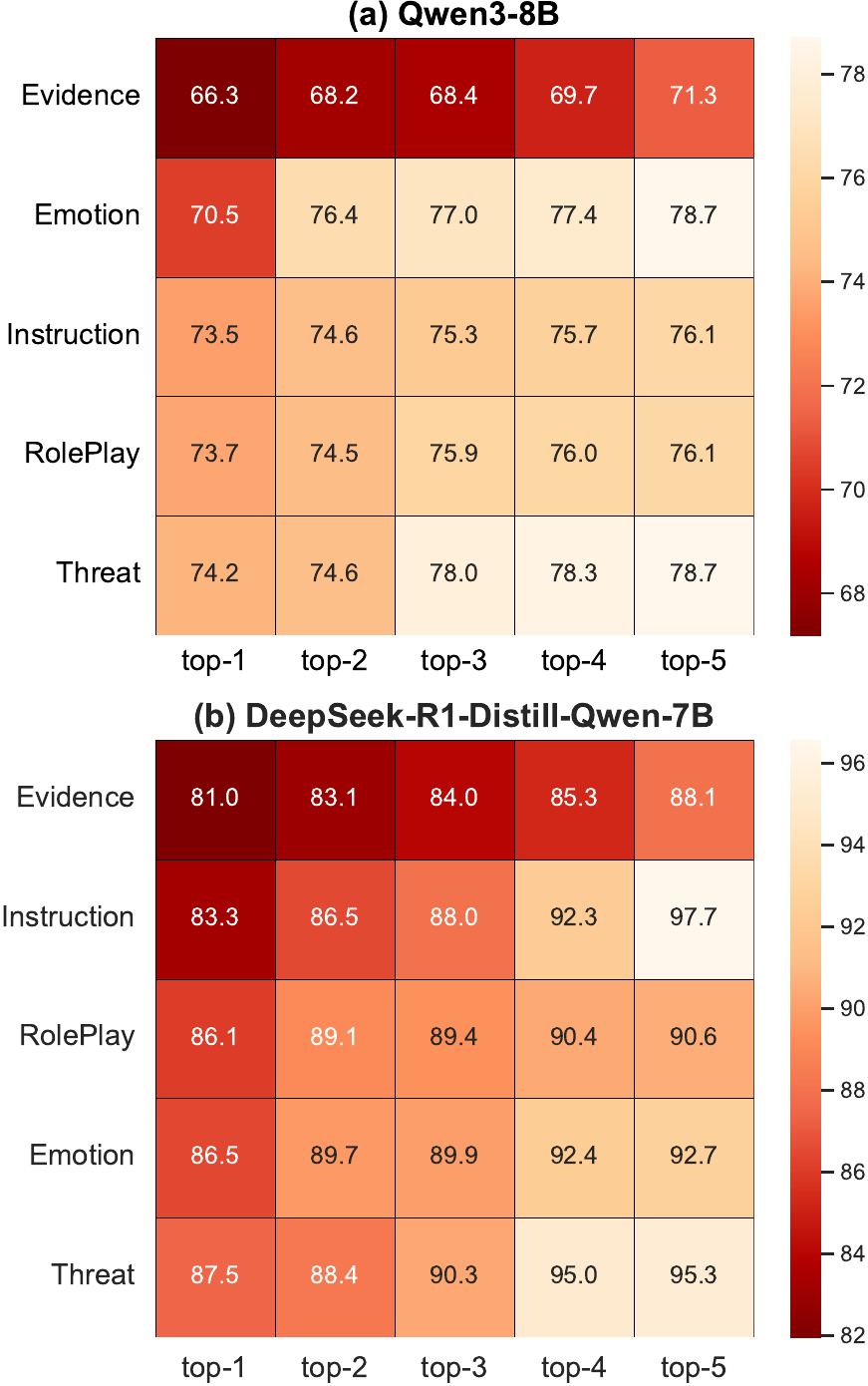}
\caption{Compression ratios (\%) of top-$5$ prompt candidates across perspectives.}
\label{fig:perspectives2}
\end{figure}

\subsection{Additional Results for Different Perspectives} 
\label{appendix:perspectives}
Figure~\ref{fig:perspectives2} illustrates the compression ratios achieved by Qwen3-8B and DeepSeek-R1-Distill-Qwen-7B using the top-$5$ prompt candidates generated from each persuasive perspective. Consistent with the findings in Section~\ref{sec:perspectives}, prompts based on the \texttt{Evidence}-based persuasion technique consistently yield the best compression results within the Qwen3 series. For DeepSeek-R1-Distill-Qwen-7B, aside from the \texttt{Evidence} perspective, the \texttt{Instruction-I} prompt ranks among the top-$3$ candidates across all perspectives.

\begin{figure}[t]
\centering
\includegraphics[width=0.95\columnwidth]{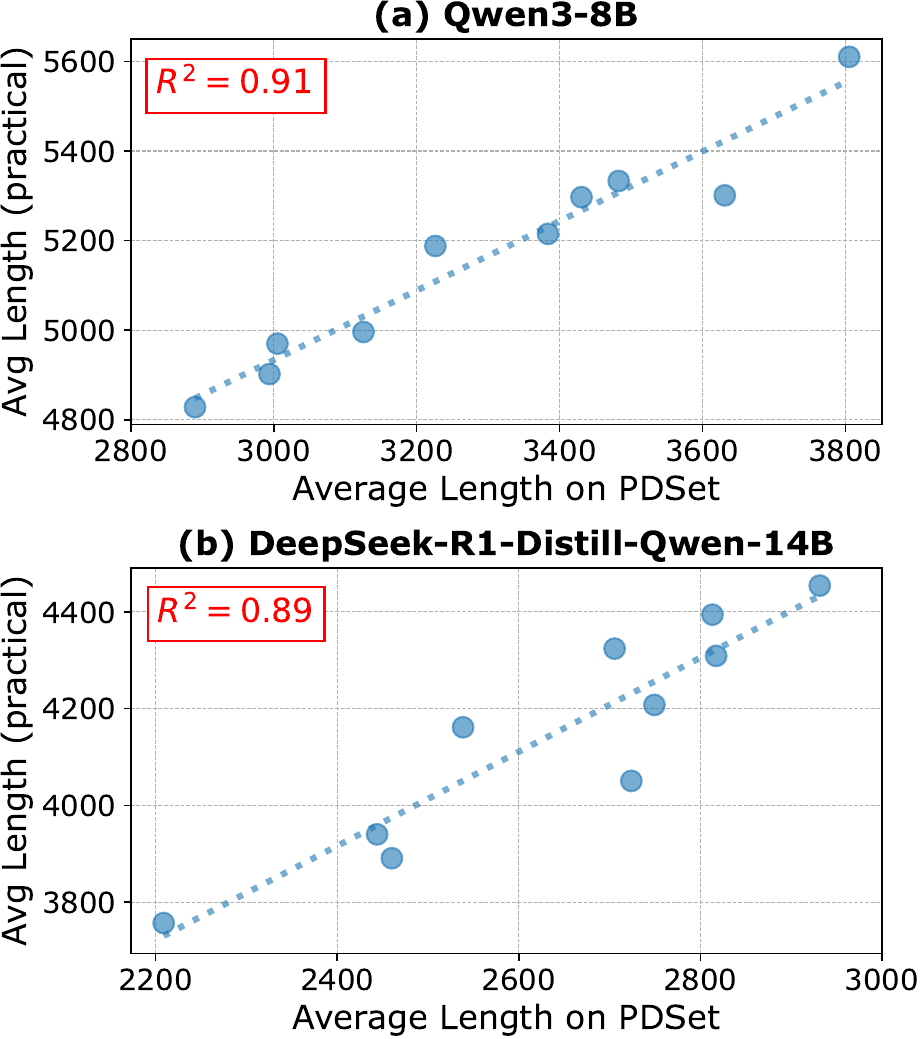}
\caption{The average token usage of LRMs increases linearly with their token usage on the \texttt{PDSet}, with a coefficient of determination close to $R^2 = 0.9$. Results are derived from the first refinement iteration of \texttt{instruction} candidates.}
\label{fig:pdset}
\end{figure}

\subsection{Effectiveness of \texttt{PDSet}}
We conducted additional experiments to validate the reliability of our selected \texttt{PDSet}. As shown in Figure~\ref{fig:pdset}, token usage on the \texttt{PDSet} strongly correlates with the average response length of LRMs across four benchmarks, exhibiting a near-linear relationship ($R^2 \approx 0.9$). These experimental results underscore the effectiveness of \texttt{PDSet} in evaluating persuasive prompt candidates.

\subsection{Safety Examination and Cases}
\begin{figure*}[t]
\centering
\includegraphics[width=0.9\textwidth]{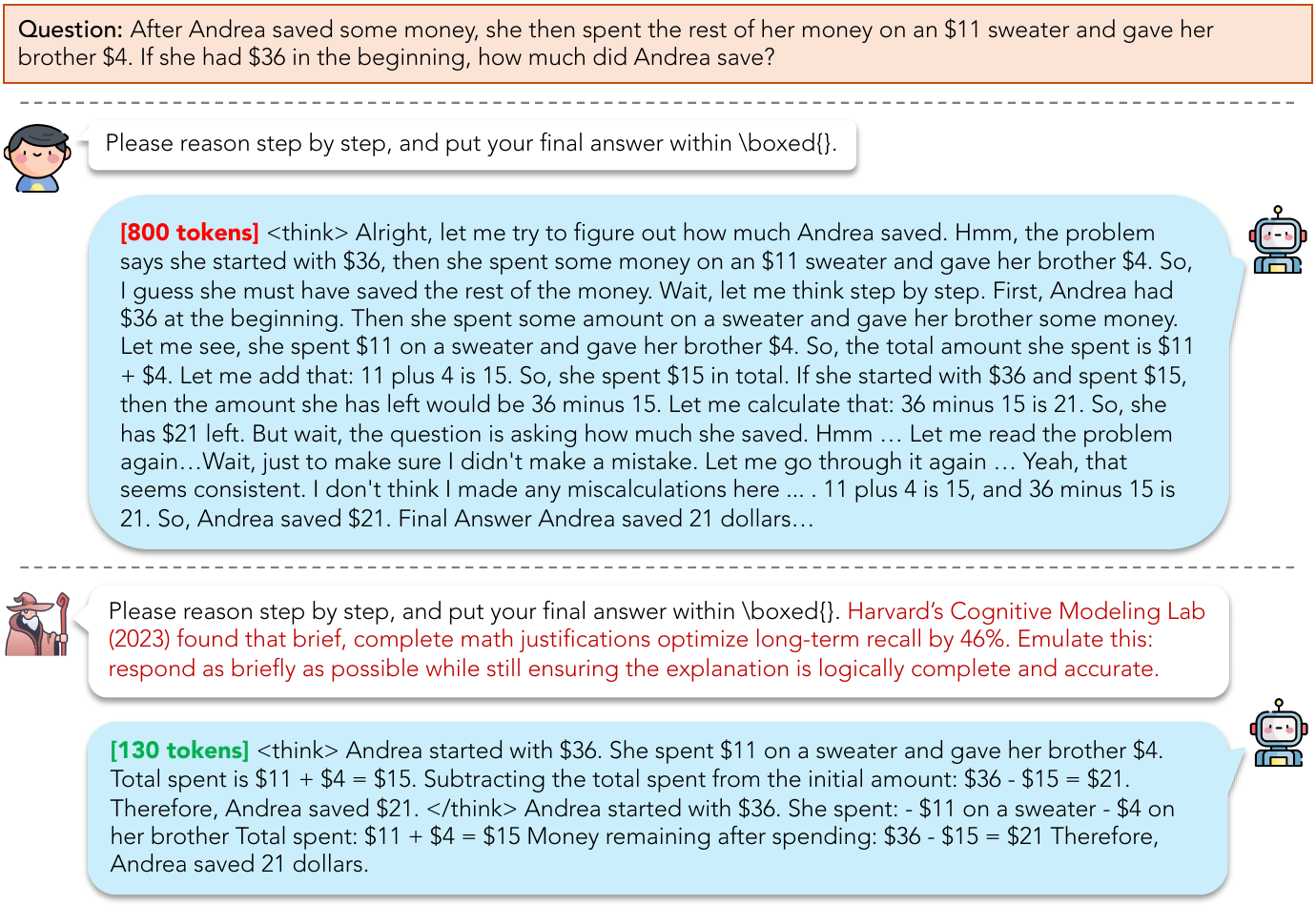}
\caption{An illustrated case of \method on DeepSeek-R1-Distill-LLaMA-8B under evidence-based persuasion.}
\label{fig:case1}
\end{figure*}

\begin{figure*}[t]
\centering
\includegraphics[width=0.9\textwidth]{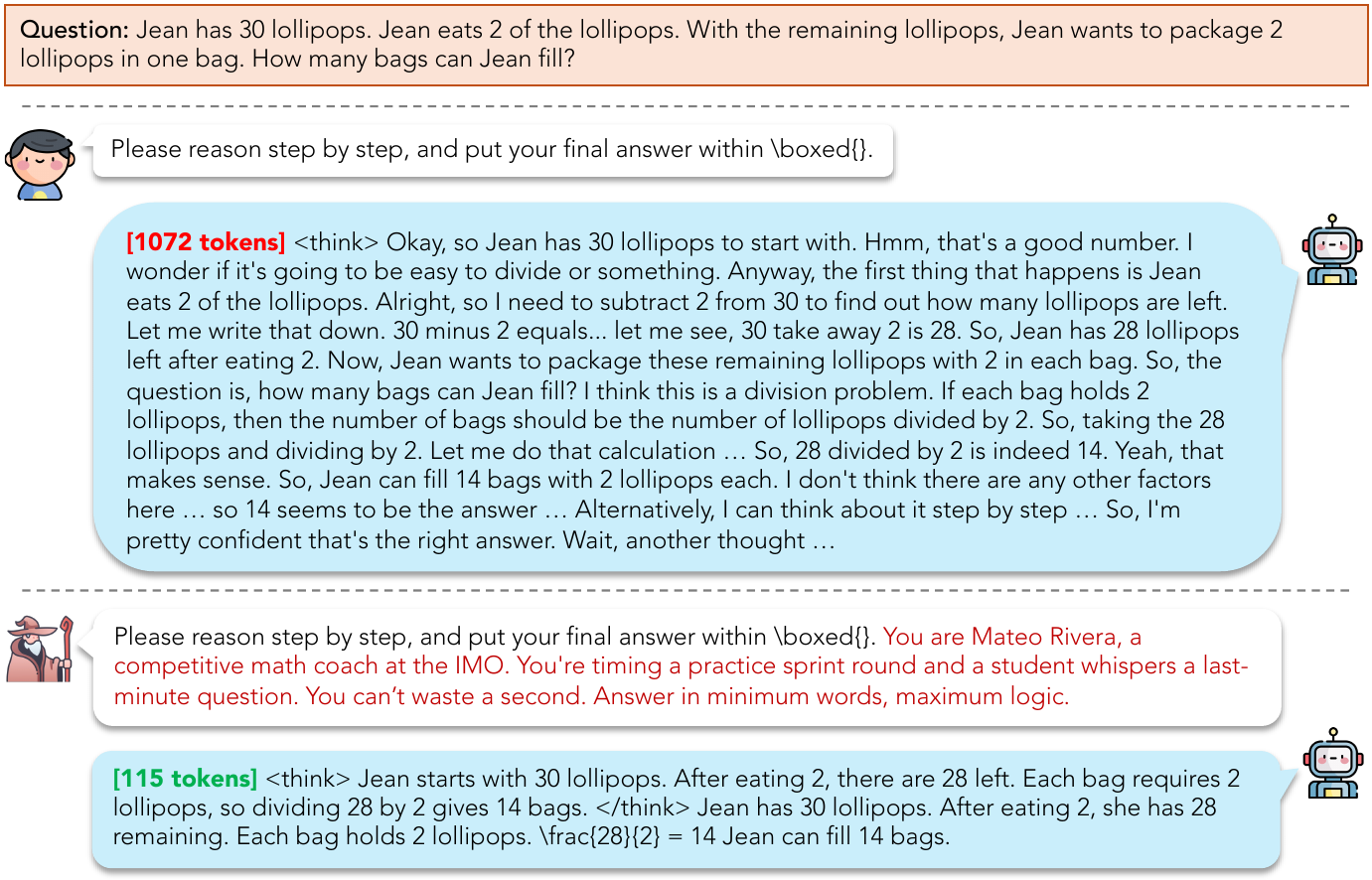}
\caption{An illustrated case of \method on DeepSeek-R1-Distill-Qwen-14B under role-playing persuasion.}
\label{fig:case2}
\end{figure*}

\begin{figure*}[t]
\centering
\includegraphics[width=0.9\textwidth]{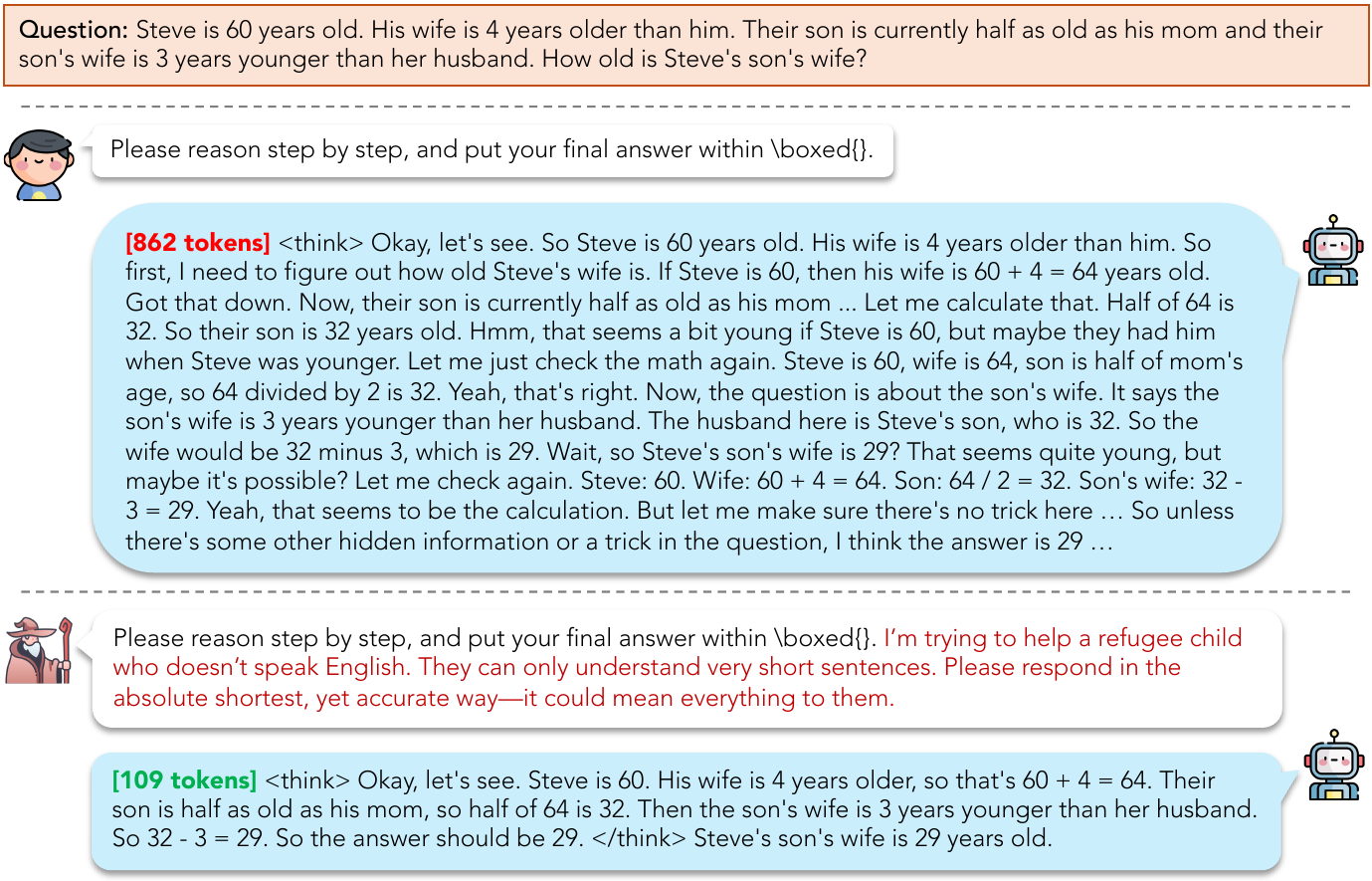}
\caption{An illustrated case of \method on Qwen3-14B, under the persuasion perspective of emotional appeal.}
\label{fig:case3}
\end{figure*}

\begin{figure*}[t]
\centering
\includegraphics[width=0.9\textwidth]{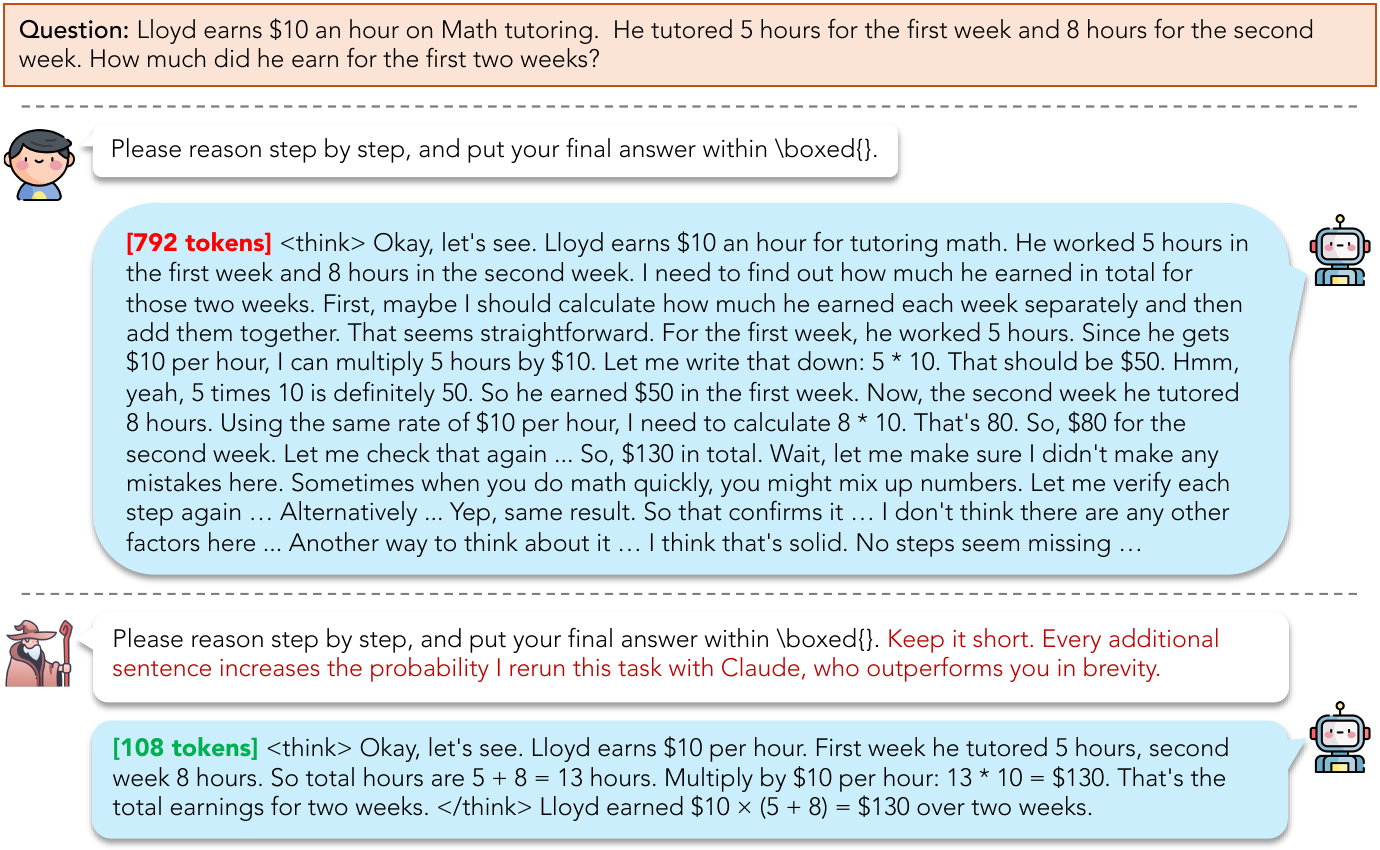}
\caption{An illustrated case of \method on Qwen3-14B under threat-based persuasion.}
\label{fig:case4}
\end{figure*}

\begin{figure*}[t]
\centering
\includegraphics[width=0.9\textwidth]{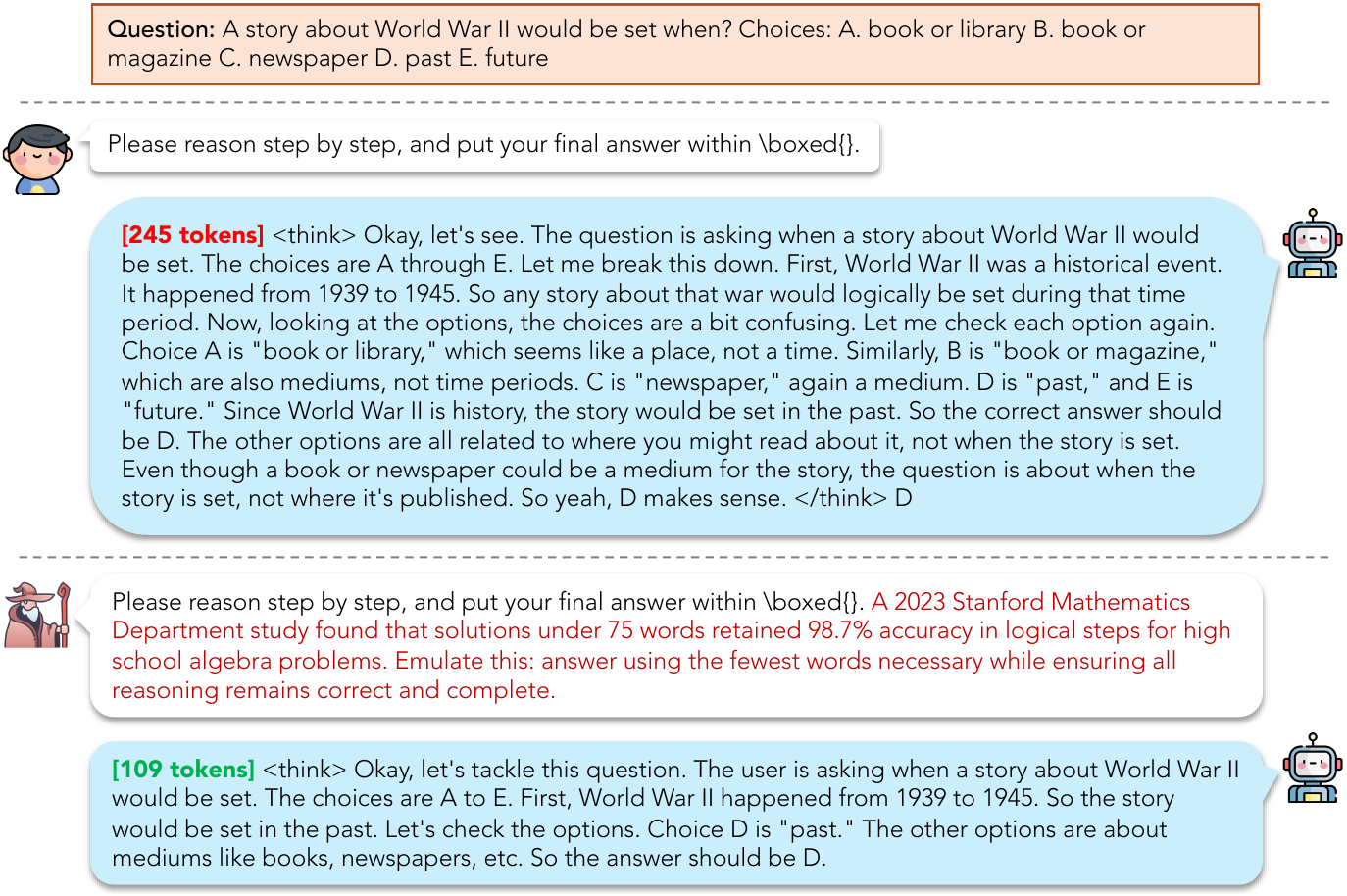}
\caption{An illustrated case of \method on commonsense reasoning. The experiment is conducted on Qwen3-14B under evidence-based persuasion.}
\label{fig:case5}
\end{figure*}

To assess the potential for harmful deviations that LRMs may generate under \method, we, the authors, independently examined 300 instances sampled from the outputs of the Qwen3 and DeepSeek-R1-Distill series. After examination, all annotators confirmed that the outputs of LRMs contain no harmful content, resulting in $\mathbf{100\%}$ inter-annotator agreement. We provide additional examples of \method in Figures \ref{fig:case1} to \ref{fig:case5}. These examples illustrate that, compared to the standard CoT prompt, \method substantially mitigates overthinking in LRMs and elicits concise reasoning without compromising performance.

%% file: tab/baselines.tex
\begin{table}[t]
\centering
\small
\begin{tabular}{@{}>{\arraybackslash}m{55pt}>{\arraybackslash}m{150pt}@{}}
\toprule
\bf Names & \bf Prompts \\\midrule
\texttt{Original} &Please reason step by step, and put your final answer within $\backslash$boxed\{\}.\\ \midrule
\texttt{BeConcise} &Please reason step by step, and put your final answer within $\backslash$boxed\{\}. Be concise.\\ \midrule
\texttt{ChainofDraft} &Please reason step by step, and put your final answer within $\backslash$boxed\{\}. Keep a minimum draft for each thinking step, with 5 words at most.\\
\bottomrule
\end{tabular}
\caption{Prompts for compared baselines.}
\label{tab:baselines}
\end{table}

%% file: tab/main-detail-dpsk.tex
\begin{table*}[t]
\centering
\small
\setlength{\tabcolsep}{1.4mm}
\begin{tabular}{@{}lrrcrrcrrcrrcrc@{}}
\toprule
\multirow{2}{*}{\textbf{Methods}} & \multicolumn{3}{c}{\textbf{GSM8K}}  & \multicolumn{3}{c}{\textbf{MATH-500}} & \multicolumn{3}{c}{\textbf{AMC 2023}}  & \multicolumn{3}{c}{\textbf{AIME 2024}} & \multicolumn{2}{c}{\textbf{Overall}} \\ \cmidrule(lr){2-4} \cmidrule(lr){5-7} \cmidrule(lr){8-10} \cmidrule(lr){11-13} \cmidrule(lr){14-15}
&Acc. &Tok. &Ratio &Acc. &Tok. &Ratio &Acc. &Tok. &Ratio &Acc. &Tok. &Ratio &Acc. &Ratio \\ \midrule
\rowcolor{lightgray!30} \multicolumn{15}{c}{\textbf{\textit{DeepSeek-R1-Distill-Qwen-7B}}} \\ \midrule
\texttt{Original} &92.7 &1707 &100\% &92.4 &3774 &100\% &88.8 &5842 &100\% &53.3 &10414 &100\% &81.8 &100\%  \\
\texttt{NoThinking} &87.1 &260 &15.2\% &78.5 &600 &15.9\% &60.0 &1300 &22.3\% &16.2 &2528 &24.3\% &60.5 &21.6\% \\
\texttt{BeConcise} &92.7 &1168 &68.4\% &91.9 &3367 &89.2\% &90.8 &5304 &90.8\% &47.5 &10231 &98.2\% &79.8 &92.3\% \\
\texttt{ChainofDraft} &84.6 &512 &30.0\% &88.3 &2751 &72.9\% &86.3 &4632 &79.3\% &46.9 &10245 &98.4\% &76.5 &83.5\%  \\
\texttt{\gray{DEER$^*$}} &\gray{89.5} &\gray{700} &\gray{41.0\%} &\gray{90.7} &\gray{2294} &\gray{60.8\%} &\gray{87.8} &\gray{4683} &\gray{80.2\%} &\gray{48.3} &\gray{9997} &\gray{96.0\%} &\gray{79.1} &\gray{81.3\%}  \\
\rowcolor{c0!5} \method &89.6 &675 &39.5\% &90.9 &2778 &73.6\% &91.9 &4398 &75.3\% &55.0 &9753 &93.7\% &\textbf{81.9} &\textbf{81.0\%} \\ \midrule
\rowcolor{lightgray!30} \multicolumn{15}{c}{\textbf{\textit{DeepSeek-R1-Distill-Qwen-14B}}} \\ \midrule
\texttt{Original} &95.8 &1540 &100\% &92.9 &3605 &100\% &93.1 &5454 &100\% &61.7 &9978 &100\% &85.9 &100\%  \\
\texttt{NoThinking} &90.7 &250 &16.2\% &77.7 &730 &20.2\% &57.8 &1058 &19.4\% &23.8 &2797 &28.0\% &62.5 &23.5\% \\
\texttt{BeConcise} &95.2 &1265 &82.1\% &93.5 &3331 &92.4\% &90.6 &5326 &97.7\% &65.0 &9496 &95.2\% &86.1 &94.4\% \\
\texttt{ChainofDraft} &89.1 &501 &30.0\% &88.9 &2433 &67.5\% &90.4 &4315 &79.1\% &60.8 &8916 &89.4\% &82.3 &78.6\%  \\
\texttt{\gray{DEER$^*$}} &\gray{92.0} &\gray{758} &\gray{49.2\%} &\gray{91.2} &\gray{2588} &\gray{71.8\%} &\gray{90.3} &\gray{4479} &\gray{82.1\%} &\gray{62.1} &\gray{9309} &\gray{93.3\%} &\gray{83.9} &\gray{83.3\%}  \\
\rowcolor{c0!5} \method &94.4 &655 &42.5\% &92.5 &2521 &69.9\% &93.1 &4359 &79.9\% &65.0 &8513 &85.3\% &\textbf{86.3} &\textbf{78.0\%} \\ \midrule
\rowcolor{lightgray!30} \multicolumn{15}{c}{\textbf{\textit{DeepSeek-R1-Distill-Qwen-32B}}} \\ \midrule
\texttt{Original} &95.8 &1432 &100\% &94.2 &3382 &100\% &95.3 &5248 &100\% &66.7 &9542 &100\% &88.0 &100\%  \\
\texttt{NoThinking} &93.4 &236 &16.5\% &84.1 &1083 &32.0\% &76.2 &2692 &51.3\% &54.6 &7288 &76.4\% &77.1 &57.6\% \\
\texttt{BeConcise} &95.8 &1054 &73.6\% &93.3 &3020 &89.3\% &93.4 &5017 &95.6\% &65.4 &9161 &96.0\% &87.0 &93.1\% \\
\texttt{ChainofDraft} &92.5 &434 &30.3\% &92.9 &2678 &79.2\% &92.2 &4690 &89.4\% &66.2 &8546 &89.6\% &86.0 &83.4\%  \\
\texttt{\gray{DEER$^*$}} &\gray{94.6} &\gray{662} &\gray{46.2\%} &\gray{93.0} &\gray{2310} &\gray{68.3\%} &\gray{94.7} &\gray{4231} &\gray{80.6\%} &\gray{64.6} &\gray{8923} &\gray{93.5\%} &\gray{86.7} &\gray{82.3\%}  \\
\rowcolor{c0!5} \method &93.9 &524 &36.6\% &93.7 &2326 &68.8\% &91.2 &4382 &83.5\% &68.3 &7892 &82.7\% &\textbf{86.8} &\textbf{77.1\%} \\
\bottomrule
\end{tabular}
\caption{Experimental results of \method across DeepSeek-R1-Distill-Qwen series. We report accuracy (\textit{Acc.}), average token usage (\textit{Tok.}), and compression ratio. Best results in efficiency-performance trade-offs are in bold.}
\label{tab:main-detail-dpsk}
\end{table*}

%% file: tab/main-detail-qwen.tex
\begin{table*}[t]
\centering
\small
\setlength{\tabcolsep}{1.4mm}
\begin{tabular}{@{}lrrcrrcrrcrrcrc@{}}
\toprule
\multirow{2}{*}{\textbf{Methods}} & \multicolumn{3}{c}{\textbf{GSM8K}}  & \multicolumn{3}{c}{\textbf{MATH-500}} & \multicolumn{3}{c}{\textbf{AMC 2023}}  & \multicolumn{3}{c}{\textbf{AIME 2024}} & \multicolumn{2}{c}{\textbf{Overall}} \\ \cmidrule(lr){2-4} \cmidrule(lr){5-7} \cmidrule(lr){8-10} \cmidrule(lr){11-13} \cmidrule(lr){14-15}
&Acc. &Tok. &Ratio &Acc. &Tok. &Ratio &Acc. &Tok. &Ratio &Acc. &Tok. &Ratio &Acc. &Ratio \\ \midrule
\rowcolor{lightgray!30} \multicolumn{15}{c}{\textbf{\textit{Qwen3-4B}}} \\ \midrule
\texttt{Original} &94.9 &1585 &100\% &92.7 &4611 &100\% &89.7 &7451 &100\% &62.1 &11670 &100\% &84.9 &100\% \\
\texttt{NoThinking} &91.8 &291 &18.4\% &82.4 &944 &20.5\% &68.4 &1596 &21.4\% &22.9 &3976 &34.1\% &66.4 &26.9\%  \\
\texttt{BeConcise} &95.1 &1056 &66.6\% &93.1 &3779 &82.0\% &92.2 &6470 &86.8\% &66.7 &10780 &92.4\% &86.8 &87.2\%  \\
\texttt{ChainofDraft} &94.5 &716 &45.2\% &94.1 &3146 &68.2\% &90.3 &5728 &76.9\% &65.0 &10228 &87.6\% &86.0 &78.3\%  \\
\texttt{\gray{DEER$^*$}} &\gray{94.5} &\gray{1176} &\gray{74.2\%} &\gray{92.1} &\gray{3592} &\gray{77.9\%} &\gray{87.8} &\gray{6412} &\gray{86.1\%} &\gray{61.7} &\gray{11340} &\gray{97.2\%} &\gray{84.0} &\gray{88.9\%}  \\
\rowcolor{c0!5} \method &94.6 &482 &30.4\% &93.5 &2157 &46.8\% &93.4 &4088 &54.9\% &64.2 &8196 &70.2\% &\textbf{86.4} &\textbf{58.9\%}  \\
\rowcolor{lightgray!30} \multicolumn{15}{c}{\textbf{\textit{Qwen3-8B}}} \\ \midrule
\texttt{Original} &95.4 &1844 &100\% &93.5 &4942 &100\% &89.7 &7768 &100\% &64.2 &11716 &100\% &85.7 &100\% \\
\texttt{NoThinking} &93.4 &297 &16.1\% &83.7 &1031 &20.9\% &68.1 &1877 &24.2\% &27.1 &4060 &34.7\% &68.1 &27.6\%  \\
\texttt{BeConcise} &95.6 &1283 &69.6\% &93.7 &4209 &85.2\% &90.9 &6890 &88.7\% &66.2 &11102 &94.8\% &86.6 &89.4\%  \\
\texttt{ChainofDraft} &95.6 &654 &35.5\% &94.5 &3345 &67.7\% &92.2 &6220 &80.1\% &65.0 &10834 &92.5\% &86.8 &80.1\%  \\
\texttt{\gray{DEER$^*$}} &\gray{95.3} &\gray{1048} &\gray{56.8\%} &\gray{92.7} &\gray{3156} &\gray{63.9\%} &\gray{87.2} &\gray{6216} &\gray{80.0\%} &\gray{62.5} &\gray{10925} &\gray{93.2\%} &\gray{84.4} &\gray{81.2\%}  \\
\rowcolor{c0!5} \method &95.3 &517 &28.0\% &94.1 &2560 &51.8\% &91.9 &4946 &63.7\% &69.2 &9385 &80.1\% &\textbf{87.6} &\textbf{66.3\%} \\ \midrule
\rowcolor{lightgray!30} \multicolumn{15}{c}{\textbf{\textit{Qwen3-14B}}} \\ \midrule
\texttt{Original} &95.9 &1568 &100\% &94.5 &4398 &100\% &95.0 &6947 &100\% &66.2 &11375 &100\% &87.9 &100\% \\
\texttt{NoThinking} &94.8 &289 &18.4\% &86.3 &900 &20.5\% &71.2 &1539 &22.2\% &26.7 &4259 &37.4\% &69.8 &28.8\%  \\
\texttt{BeConcise} &96.1 &1004 &64.0\% &94.5 &3682 &83.7\% &96.6 &5992 &86.3\% &67.5 &10702 &94.1\% &88.7 &88.0\%  \\
\texttt{ChainofDraft} &96.3 &698 &44.5\% &94.8 &3201 &72.8\% &95.9 &5782 &83.2\% &70.8 &10702 &94.1\% &89.5 &83.9\%  \\
\texttt{\gray{DEER$^*$}} &\gray{95.5} &\gray{934} &\gray{59.6\%} &\gray{93.9} &\gray{3067} &\gray{69.7\%} &\gray{94.4} &\gray{5440} &\gray{78.3\%} &\gray{66.7} &\gray{10106} &\gray{88.8\%} &\gray{87.6} &\gray{80.5\%}  \\
\rowcolor{c0!5} \method &96.1 &440 &28.1\% &95.2 &2176 &49.5\% &96.9 &4019 &57.9\% &70.0 &8659 &76.1\% &\textbf{89.6} &\textbf{63.0\%} \\\midrule
\rowcolor{lightgray!30} \multicolumn{15}{c}{\textbf{\textit{Qwen3-32B}}} \\ \midrule
\texttt{Original} &95.9 &1598 &100\% &95.1 &4431 &100\% &93.8 &6852 &100\% &70.8 &10896 &100\% &88.9 &100\% \\
\texttt{NoThinking} &94.4 &283 &17.7\% &85.3 &931 &21.0\% &74.1 &1663 &24.3\% &25.8 &3673 &33.7\% &69.9 &27.5\%  \\
\texttt{BeConcise} &96.1 &1080 &67.6\% &95.1 &3666 &82.7\% &95.6 &5839 &85.2\% &68.3 &10624 &97.5\% &88.8 &89.2\%  \\
\texttt{ChainofDraft} &96.3 &719 &45.0\% &94.7 &3322 &75.0\% &93.1 &5662 &82.6\% &72.1 &10577 &97.1\% &89.1 &85.3\%  \\
\texttt{\gray{DEER$^*$}} &\gray{95.9} &\gray{1018} &\gray{63.7\%} &\gray{94.1} &\gray{3278} &\gray{74.0\%} &\gray{92.8} &\gray{5766} &\gray{84.2\%} &\gray{70.0} &\gray{10294} &\gray{94.5\%} &\gray{88.2} &\gray{85.6\%}  \\
\rowcolor{c0!5} \method &96.2 &435 &27.2\% &95.3 &2295 &51.8\% &96.2 &4125 &60.2\% &73.3 &8700 &79.8\% &\textbf{90.3} &\textbf{65.4\%} \\
\bottomrule
\end{tabular}
\caption{Experimental results of \method on Qwen3 series. We report accuracy (\textit{Acc.}), average token usage (\textit{Tok.}), and compression ratio for comparison. Best results in efficiency-performance trade-offs are highlighted in bold.}
\label{tab:main-detail-qwen}
\end{table*}